%% file: arxiv.tex
\documentclass{article}
\usepackage{iclr2027_conference,times}
\input{math_commands.tex}

\usepackage{amsmath,amssymb,amsthm,mathtools,bm}
\usepackage{booktabs,tabularx}
\usepackage{enumitem}
\usepackage{microtype}
\usepackage{algorithm}
\usepackage{algorithmic}
\usepackage{hyperref}
\usepackage{url}
\usepackage{tikz}
\usetikzlibrary{decorations.pathreplacing}
\hypersetup{
  hidelinks,
  pdftitle={Stochastic Tracking of Drifting Affine Fixed Points},
  pdfauthor={Sajad Khodadadian}
}
\newtheorem{theorem}{Theorem}
\newtheorem{lemma}[theorem]{Lemma}
\newtheorem{proposition}[theorem]{Proposition}
\newtheorem{corollary}[theorem]{Corollary}
\newtheorem{assumption}[theorem]{Assumption}
\theoremstyle{remark}

\newcommand{\Pp}{\mathbb{P}}
\newcommand{\cF}{\mathcal{F}}
\newcommand{\cG}{\mathcal{G}}
\newcommand{\cX}{\mathcal{X}}
\newcommand{\cY}{\mathcal{Y}}
\newcommand{\cE}{\mathcal{E}}
\newcommand{\cS}{\mathcal{S}}
\newcommand{\cT}{\mathcal{T}}
\newcommand{\op}{\mathrm{op}}
\newcommand{\ip}[2]{\left\langle #1,#2\right\rangle}
\newcommand{\norm}[1]{\left\lVert #1\right\rVert}
\newcommand{\abs}[1]{\left\lvert #1\right\rvert}
\newcommand{\ind}{\mathbf{1}}
\newcolumntype{L}[1]{>{\raggedright\arraybackslash}p{#1}}

\title{Stochastic Tracking of Drifting Affine Fixed Points}
\author{Sajad Khodadadian\\
Grado Department of Industrial and Systems Engineering\\
Virginia Tech}

 \iclrfinalcopy
\begin{document}
\raggedbottom
\maketitle

\begin{abstract}
Linear stochastic approximation (SA) is widely used to estimate fixed points from noisy observations, but most finite-time analyses assume that the underlying operator remains fixed. We study linear SA with time-varying affine operators, where the fixed points change over time and the observation noise may be correlated across rounds. For constant-stepsize updates, we establish a bound on the cumulative expected squared distance between the iterates and the corresponding fixed points. The bound accounts for the initial error, observation noise, and changes in the fixed points over time. We also establish lower bounds for scalar Gaussian models, showing that, under specified conditions, a suitably chosen constant stepsize achieves the smallest possible worst-case cumulative error up to constant factors, even compared with estimators that use future observations. These results further quantify the effect of temporal correlation on estimation accuracy. Since choosing an appropriate stepsize generally requires knowledge of the noise level and how much the fixed points change, we develop an adaptive method that uses an additional bounded, unbiased observation of the fixed point to achieve an almost optimal algorithm without needing the knowledge of the noise level and the fixed point variation over time. Finally, we apply our results to value-function estimation using synchronous tabular temporal-difference learning in changing environments, assuming generative-model access and a common invariant distribution.
\end{abstract}

\section{Introduction}

Reinforcement learning algorithms often operate in environments whose conditions change over time. A navigation system may encounter changing traffic conditions, while a recommendation system may face shifts in user preferences. In such settings, the rewards and consequences of an action can change, making earlier observations less representative of the decisions the learner currently faces. These challenges have motivated a growing literature on reinforcement learning in nonstationary environments~\citep{cheung2020reinforcement,wei2021nonstationary}.

A basic part of this problem is keeping value estimates up to date. Policy evaluation estimates the expected future reward obtained by following a given policy, and these estimates support policy improvement in many reinforcement learning algorithms. Finite-time analyses of temporal-difference (TD) learning typically study a fixed policy in a fixed environment~\citep{bhandari2018finite,khamaru2021td}. When rewards or transition probabilities change, however, the value function changes as well. Even when the environment remains fixed, changing the policy can change its value function and a critic that evaluates the current policy must keep its estimates up to date as the policy changes~\citep{walton2020adiabatic}. The learner must therefore maintain accurate estimates throughout learning, while the quantity being estimated continues to move.

Tracking a changing value function requires balancing responsiveness and noise reduction. A small step size reduces the influence of individual noisy observations but can leave the estimate slow to respond to change. A large step size responds more quickly but produces larger fluctuations. Correlation between observations adds another difficulty: when errors persist across successive samples, averaging them may reduce noise more slowly than it would with independent samples. A useful analysis must account for both the movement of the target and the information available in the observations.

This tradeoff is well established in stochastic tracking and time-varying optimization~\citep{bastianello2024stochastic,cutler2023distributional,sahu2026momentum}. For exact contractive updates, \citet{hadiji2026tracking} bound cumulative squared tracking error using the sum of squared changes in the target. This measure captures the difference between gradual movement and abrupt changes, even when the overall displacement is the same. We build on this deterministic result to study how temporally correlated observation noise affects the tracking guarantee.

We consider a deterministic sequence of affine contractions $F_t(x)=M_tx+b_t$, each with its own fixed point. At each round, the learner observes a noisy evaluation of the current map and updates its estimate using a constant step size. This formulation includes discounted Bellman updates under the conditions of our TD application. The affine structure allows us to separate the error caused by initialization and target movement from the error caused by noise. We use this separation to obtain a cumulative mean-square bound that retains the deterministic tracking terms and makes the contribution of temporal correlation explicit.

\paragraph{Our contributions.}
\begin{enumerate}[leftmargin=*,itemsep=1pt,topsep=2pt]

\item \textbf{A bound on tracking error.}
We bound the total expected squared distance between the learner's estimates and the changing fixed points. The bound separates the contributions of the initial error, observation noise, and target movement. It shows how the step size and correlation between observations affect tracking accuracy, extending an existing guarantee for exact updates to noisy observations (Theorem~\ref{thm:upper}).

\item \textbf{Limits on tracking accuracy.}
For a scalar model with Gaussian noise, we establish a worst-case lower bound on the error of any estimator, even one that can use future observations. Over specified ranges of target movement, this lower bound matches, up to constant factors, the guarantee achieved by a suitably chosen step size. We also establish how temporal correlation increases this unavoidable error (Theorems~\ref{thm:lower} and~\ref{thm:correlated-lower}).

\item \textbf{Combining trackers with different step sizes.}
The appropriate step size depends on the noise level and target movement, which may be unknown. We therefore run several trackers with different step sizes and combine their predictions. To compare their accuracy, the method uses additional bounded observations that are unbiased estimates of the current target. The combined prediction has total expected squared error at most that of the best tracker in the collection plus an explicit penalty that grows logarithmically with the number of trackers (Theorem~\ref{thm:meta}).

\item \textbf{Application to policy evaluation.}
We apply the tracking bound to synchronous tabular TD with changing
rewards and transition probabilities, assuming simulator access and
a common invariant distribution. Appendix~\ref{app:td} gives the
tracking guarantees, constructs the additional target observations
needed for adaptation, and quantifies their sampling cost. We also
extend the analysis to coupled linear updates under explicit
contraction assumptions (Appendix~\ref{app:ttsa}).
\end{enumerate}

Section~\ref{sec:related} discusses prior work, Section~\ref{sec:model} gives the problem formulation, and Sections~\ref{sec:results}--\ref{sec:discussion} present the results and their scope. Complete proofs appear in the appendix.

\section{Related work}
\label{sec:related}

\paragraph{Tracking and dynamic regret.}
\citet{besbes2015nonstationary} establish variation-dependent minimax guarantees for nonstationary stochastic optimization, using a budget on changes in the objective. \citet{yang2016tracking} study moving minimizers under exact and noisy gradient feedback, while \citet{zhang2017improved} introduce squared path length into dynamic-regret bounds with multiple gradient queries. For contractive updates, \citet{bernstein2018asynchronous} analyze asynchronous tracking. \citet{madden2021tracking} establish asymptotic tracking bounds and lower bounds for first-order methods in time-varying optimization, while \citet{hadiji2026tracking} bound cumulative squared error by quadratic path variation. With noisy feedback, \citet{bastianello2024stochastic} treat general time-varying operators, and \citet{cutler2023distributional} separate optimization, noise, and minimizer drift. \citet{sahu2026momentum} study the same stepsize tradeoff and the limitations of momentum in nonstationary optimization. Section~\ref{sec:discussion} compares the assumptions and guarantees with our affine result.

\paragraph{Sequence estimation and adaptation.}
The scalar class $\sum_{t<T}(\theta_{t+1}-\theta_t)^2\leq W$ is a discrete Sobolev class. \citet{sadhanala2016total} give its minimax estimation bounds. \citet[Section~3.5]{baby2019online} obtain online forecasting guarantees that adapt to unknown variation, with the same leading rate up to logarithmic factors and additional terms. Their method assumes a known horizon and sub-Gaussian noise parameter, uses noisy scalar target observations, and requires no separate validation stream. \citet{baby2021optimal} develop variation-adaptive guarantees for exp-concave online learning, including squared loss. Correlated errors in nonparametric regression are also well studied~\citep{opsomer2001correlated}. Our Gaussian results specialize this setting to affine queries and keep the finite-horizon dependence on the covariance factor explicit. For general affine maps, we assume a target-validation oracle and apply standard exponential weighting~\citep{cesabianchi2006prediction} to tracking recursions.

\paragraph{Stochastic approximation.}
\citet{konda2003slow} study slowly changing Markov statistics with decreasing stepsizes, and \citet{kumar2019nonasymptotic} analyze constant-stepsize tracking. For fixed targets, finite-sample analyses cover general norms, multiplicative noise, and high-probability guarantees~\citep{chen2020contractive,chandak2026contractive,diakonikolas2026highprob}. Related two-timescale results are discussed alongside the extension in Appendix~\ref{app:ttsa}.

\paragraph{Policy evaluation and nonstationary RL.}
Finite-time bounds for stationary TD are established
by \citet{bhandari2018finite}. In an instance-dependent
analysis of tabular policy evaluation,
\citet{khamaru2021td} show that standard TD can be
suboptimal and establish near-optimal guarantees
for variance-reduced stochastic approximation
methods. \citet{totaro2021fast} propose a
Kalman-style approach to changing value parameters. Most nonstationary RL theory concerns control and exploration~\citep{cheung2020reinforcement,mao2021near,wei2021nonstationary,zhou2022nonstationary,feng2023nonstationary,jali2026natural}. Closer to our application, \citet{walton2020adiabatic} analyzes tabular TD and Q-learning along a changing Markov trajectory. Our TD guarantee concerns cumulative estimation error under synchronous generative access.

\section{Problem formulation}
\label{sec:model}

We study the problem of tracking a changing fixed point using noisy observations. Fix a horizon $T\geq2$. At each time $t=1,\ldots,T$, consider an affine map
\[
F_t(x)=M_tx+b_t,
\qquad
M_t\in\R^{d\times d},
\quad
b_t\in\R^d.
\]
The sequence of maps is deterministic and may change over time. We equip $\R^d$ with a fixed inner product
$\ip{\cdot}{\cdot}$ and its associated norm
$\norm{x}=\sqrt{\ip{x}{x}}$. We write
$\norm{\cdot}_{\op}$ for the induced operator norm.
\begin{assumption}[Uniform contraction]
\label{ass:contraction}
There is a constant $\kappa\in(0,1]$ such that
$\norm{M_t}_{\op}\leq1-\kappa$ for every $t=1,\ldots,T$.
\end{assumption}

Under this assumption, each map has a unique fixed point,
\[
x_t^*=F_t(x_t^*)=(I-M_t)^{-1}b_t.
\]
The learner's task is to estimate $x_t^*$ at each time $t$ using the observations available before that time.

\paragraph{Observations and updates.}
The learner starts from a deterministic initial estimate $x_1\in\R^d$. Let $\cF_t$ denote the information available before the observation at time $t$, including past observations and any internal randomness used by the learner. The estimate $x_t$ is $\cF_t$-measurable. After forming this estimate, the learner receives a noisy evaluation of the current map:
\[
\widehat F_t(x_t)=F_t(x_t)+\xi_{t+1},
\qquad t=1,\ldots,T-1,
\]
where the observation error $\xi_{t+1}$ is $\cF_{t+1}$-measurable.

We analyze the update
\begin{align}    
x_{t+1}
=(1-\eta)x_t+\eta\widehat F_t(x_t),
\qquad t=1,\ldots,T-1,\label{eq:SA_rec}
\end{align}
with a constant stepsize $\eta\in(0,1]$. Thus, each new estimate combines the previous estimate with the latest noisy evaluation. 

\paragraph{Noise assumptions.}
We allow observation errors to be correlated across time. In addition to requiring each error to have mean zero and bounded second moment, we control how much noise can accumulate in a linear combination of observations.

\begin{assumption}[Noise and temporal correlation]
\label{ass:covariance}
There are constants $\sigma^2\geq0$ and $\tau\geq1$ such that, for every $t=1,\ldots,T-1$,
\[
\E[\xi_{t+1}]=0,
\qquad
\E\norm{\xi_{t+1}}^2\leq\sigma^2.
\]
Moreover, for every integer $1\leq n\leq T-1$ and every collection of deterministic matrices
$C_1,\ldots,C_n\in\R^{d\times d}$,
\begin{equation}
\E\norm{\sum_{s=1}^n C_s\xi_{s+1}}^2
\leq
\tau\sigma^2
\sum_{s=1}^n\norm{C_s}_{\op}^2.
\label{eq:integrated-covariance}
\end{equation}
\end{assumption}

Here, $\sigma^2$ bounds the size of an individual observation error, while $\tau$ controls the effect of temporal correlation on accumulated noise. A standard special case is martingale-difference noise:
\[
\E[\xi_{t+1}\mid\cF_t]=0,
\qquad
\E[\norm{\xi_{t+1}}^2\mid\cF_t]\leq\sigma^2.
\]
These conditions imply Assumption~\ref{ass:covariance} with $\tau=1$. The general assumption requires only unconditional mean-zero errors and therefore also permits correlated noise. For example, centered scalar noise with
\[
\E[\xi_{s+1}\xi_{t+1}]
=\sigma^2\chi^{|s-t|},
\qquad 0\leq\chi<1,
\]
satisfies the assumption with $\tau=(1+\chi)/(1-\chi)$. Proposition~\ref{prop:geometric-correlation} gives the corresponding matrix result.

\paragraph{Tracking objective.}
We measure performance by the cumulative expected squared distance between the learner's estimates and the current fixed points:
\[
\cE_T
:=\sum_{t=1}^{T}\E\norm{x_t-x_t^*}^2.
\]
To describe how much the target changes, we use its squared variation,
\[
W_T
:=\sum_{t=1}^{T-1}\norm{x_{t+1}^*-x_t^*}^2.
\]
This quantity captures the magnitudes of successive changes and
distinguishes movement concentrated in a few rounds from movement
spread more evenly over time. For example, a displacement $v\in\R^d$ made in one jump contributes $\norm{v}^2$, whereas the same displacement spread evenly over $T-1$ steps contributes $\norm{v}^2/(T-1)$.

Our goal is to bound $\cE_T$ in terms of the initial error, the contraction strength, the noise, and the target variation, and to understand how the stepsize affects tracking accuracy.

\section{Main results}
\label{sec:results}

\subsection{Finite-time tracking bound}

We first bound the cumulative tracking error for any constant stepsize. The result quantifies the effects of initialization, observation noise, and movement of the fixed points.

\begin{theorem}[Cumulative tracking error]
\label{thm:upper}
Suppose Assumptions~\ref{ass:contraction}--\ref{ass:covariance} hold. For every constant stepsize $\eta\in(0,1]$,
\begin{equation}
\boxed{
\cE_T
\leq
\frac{\norm{x_1-x_1^*}^2}{\eta\kappa}
+\frac{\eta\tau\sigma^2T}{\kappa(2-\eta\kappa)}
+\frac{W_T}{\eta^2\kappa^2}.}
\label{eq:upper-main}
\end{equation}
\end{theorem}

The first term bounds the contribution of the initial estimation error. The second accounts for accumulated observation noise, while the third accounts for the lag behind the moving target. Increasing the stepsize reduces the initialization and movement terms but increases the noise term. The covariance factor $\tau$ enters only through the noise term.

To obtain an explicit stepsize choice, we next assume exact initialization, $x_1=x_1^*$. This removes the first term and isolates the tradeoff between noise and target movement. Using $2-\eta\kappa\geq1$, we obtain the simpler bound
\[
\cE_T
\leq
\frac{\tau\sigma^2T}{\kappa}\eta
+\frac{W_T}{\kappa^2}\eta^{-2}.
\]
Minimizing this expression over positive stepsizes gives the following corollary, provided the minimizing stepsize does not exceed one.

\begin{corollary}[Tracking bound with a tuned stepsize]
\label{cor:tuned}
Under the assumptions of Theorem~\ref{thm:upper}, suppose
$x_1=x_1^*$, $W_T>0$, and $\tau\sigma^2>0$. Define $\eta_{\rm bal}
:=
\left(
\frac{2W_T}{\kappa\tau\sigma^2T}
\right)^{1/3}$. If $\eta_{\rm bal}\leq1$, running the iteration with
$\eta=\eta_{\rm bal}$ gives
\[
\cE_T
<
2\frac{(\tau\sigma^2T)^{2/3}W_T^{1/3}}{\kappa^{4/3}}.
\]
\end{corollary}

The tuned stepsize increases with target variation and decreases with the noise level and covariance factor. This choice requires knowledge of those quantities and minimizes the simplified upper bound above. Theorem~\ref{thm:upper} itself requires neither exact initialization nor positive noise or variation, and remains valid for every $\eta\in(0,1]$.

\subsection{A scalar Gaussian benchmark}
\label{sec:gaussian}

We next examine whether the tracking rate in Corollary~\ref{cor:tuned} can be improved when the observation noise is independent. We consider a scalar model in which noisy operator evaluations can be converted into direct observations of the changing target.

Consider the affine maps
\[
F_t^\theta(x)=(1-\kappa)x+\kappa\theta_t,
\]
whose fixed points are $\theta_t$. These are also the Bellman maps of a one-state reward process with discount factor $1-\kappa$ and mean reward $\kappa\theta_t$.

At each round, the learner may make at most $q$ queries, where $q\geq1$ is an integer. Query locations may depend on previous observations. The response to the $i$th query at location $x$ is
\[
\mathsf O_{t,i}(x)
=F_t^\theta(x)+\varepsilon_{t,i},
\]
where the errors are independent across queries and rounds, with
$\varepsilon_{t,i}\sim\mathcal N(0,\sigma^2)$ and $\sigma^2>0$.

For a variation budget $W>0$, define the target class
\[
\Theta_T(W)
:=
\left\{
\theta\in\R^T:
\theta_1=0,\quad
\sum_{t=1}^{T-1}(\theta_{t+1}-\theta_t)^2\leq W
\right\}.
\]
Thus, $W$ bounds the variation of every sequence in the class, whereas $W_T$ denotes the variation of an individual sequence. The initial target $\theta_1=0$ is known to the learner.

When $\kappa$ is known, each response can be transformed as
\[
\frac{\mathsf O_{t,i}(x)-(1-\kappa)x}{\kappa}
=
\theta_t+\frac{\varepsilon_{t,i}}{\kappa}.
\]
The query location disappears from this expression. Averaging $q$ transformed responses therefore gives a Gaussian observation of $\theta_t$ with variance
\[
\nu^2:=\frac{\sigma^2}{q\kappa^2}.
\]
The problem reduces to estimating a sequence from noisy observations under a bound on its squared increments. Writing
$(D\theta)_t=\theta_{t+1}-\theta_t$, this constraint is
$\norm{D\theta}_2^2\leq W$, commonly called a discrete Sobolev constraint. Its estimation rates are classical~\citep{sadhanala2016total}. The following theorem specializes the lower bound to our query model, with an explicit variation range and a fixed initial target.

\begin{theorem}[Gaussian lower bound for moderate variation]
\label{thm:lower}
In the scalar model above, suppose the learner is told
$\kappa$, $\sigma^2$, $q$, $W$, and $\theta_1=0$.
Let $\nu^2=\sigma^2/(q\kappa^2)$.
There is a universal constant $c>0$ such that, for $T\geq16$ and
\[
\frac{16\nu^2}{T^2}
\leq W
\leq
\frac{\nu^2T}{32},
\]
\[
\inf_{\widehat\theta}
\sup_{\theta\in\Theta_T(W)}
\E_\theta\sum_{t=1}^{T}
(\widehat\theta_t-\theta_t)^2
\geq
c\frac{(\sigma^2T/q)^{2/3}W^{1/3}}{\kappa^{4/3}}.
\]
The infimum allows estimators to use all $qT$ observations before producing any estimate. The bound therefore also applies to online algorithms, which have access only to past observations.
\end{theorem}

The appendix gives a self-contained proof with $c=1/4096$. We now compare this lower bound with the constant-stepsize tracker. Averaging $q$ independent responses at each update reduces the oracle noise variance to $\sigma^2/q$, allowing us to apply Theorem~\ref{thm:upper} with this smaller variance.

\begin{corollary}[Matching tracking guarantee]
\label{cor:overlap}
In the same scalar model, start from $x_1=0$. At each update time $t=1,\ldots,T-1$, make $q$ queries at the current iterate and average the responses. Define
\[
\eta_{q,\rm bal}
:=
\left(
\frac{2qW}{\kappa\sigma^2T}
\right)^{1/3}.
\]
If $\eta_{q,\rm bal}\leq1$, running the iteration in \eqref{eq:SA_rec} with
$\eta=\eta_{q,\rm bal}$ gives
\[
\sup_{\theta\in\Theta_T(W)}
\E_\theta\sum_{t=1}^{T}(x_t-\theta_t)^2
<
2\frac{(\sigma^2T/q)^{2/3}W^{1/3}}{\kappa^{4/3}}.
\]
Consequently, for $T\geq16$, the upper and lower bounds agree up to universal constants whenever
\[
\frac{16\nu^2}{T^2}
\leq W
\leq
\min\left\{
\frac{\nu^2T}{32},
\frac{\kappa^3\nu^2T}{2}
\right\}.
\]
\end{corollary}

Within this range, the constant-stepsize tracker attains the smallest possible worst-case error up to constants, despite using only past observations. The two upper restrictions on $W$ come from the lower-bound theorem and the requirement that the balancing stepsize be at most one.

\subsection{The cost of temporal correlation}
\label{sec:correlated-lower}

The previous subsection considered independent observation errors. We now allow errors to be correlated across rounds and show that the dependence on $\tau$ in the tuned tracking bound is necessary over a specified range of target variation.

We retain the scalar maps $F_t^\theta$ and target class $\Theta_T(W)$ from Section~\ref{sec:gaussian}. The $i$th query at round $t$ uses error $\varepsilon_{t,i}$. For each $i=1,\ldots,q$, the sequence $(\varepsilon_{t,i})_{t=1}^T$ is jointly Gaussian, with
\[
\E[\varepsilon_{t,i}]=0,
\qquad
\E[\varepsilon_{s,i}\varepsilon_{t,i}]
=\sigma^2\chi^{|s-t|},
\qquad
\chi=\frac{\tau-1}{\tau+1}.
\]
Sequences corresponding to different query indices are independent. Thus, $\tau=1$ gives independent errors, while larger values of $\tau$ give more persistent temporal correlation. Averaging the $q$ responses at each round produces noise satisfying Assumption~\ref{ass:covariance} with variance parameter $\sigma^2/q$ and the same covariance factor $\tau$.

\begin{theorem}[Gaussian lower bound with temporal correlation]
\label{thm:correlated-lower}
Let $T\geq16$, $\sigma^2>0$, $q\geq1$ be an integer,
$1\leq\tau\leq T/8$, and
$\nu^2=\sigma^2/(q\kappa^2)$.
For the correlated observations above, if
\[
\frac{4\tau\nu^2}{T^2}
\leq W
\leq
\frac{\nu^2T}{128\tau^2},
\]
then
\[
\inf_{\widehat\theta}
\sup_{\theta\in\Theta_T(W)}
\E_\theta\sum_{t=1}^T
(\widehat\theta_t-\theta_t)^2
\geq
\frac{1}{8192}
\frac{(\tau\sigma^2T/q)^{2/3}W^{1/3}}{\kappa^{4/3}}.
\]
The bound holds even if the learner knows
$\kappa,\sigma^2,q,W,\tau$, and $\theta_1=0$, and receives all $qT$ observations before estimating the targets.
\end{theorem}

To compare this lower bound with the tracker, start from $x_1=0$, average $q$ responses at each of the first $T-1$ update times, and choose
\[
\eta
=
\left(
\frac{2qW}{\kappa\tau\sigma^2T}
\right)^{1/3}.
\]
If $W\leq\kappa^3\tau\nu^2T/2$, this stepsize is at most one. Theorem~\ref{thm:upper} then bounds the worst-case cumulative error over $\Theta_T(W)$ by a quantity smaller than
$2(\tau\nu^2T)^{2/3}W^{1/3}$.
Consequently, the upper and lower bounds match up to universal constants whenever
\[
\frac{4\tau\nu^2}{T^2}
\leq W
\leq
\min\left\{
\frac{\nu^2T}{128\tau^2},
\frac{\kappa^3\tau\nu^2T}{2}
\right\}.
\]
Under the theorem's conditions, this interval is nonempty when $\kappa T\geq2$. Within this range, the factor $\tau^{2/3}$ is unavoidable in the Gaussian model, even when the noise covariance is known. Appendix~\ref{app:correlated-lower} proves the lower bound using target sequences that change slowly relative to the decay of noise correlations.

\paragraph{A constant-target example.}
To isolate the effect of temporal correlation, consider a target that
remains fixed at zero: $\theta_t=0$ for every $t\geq1$.
Initialize the tracker at $x_1=0$, so there is no initial error or target
movement, and all subsequent estimation error comes from observation noise.
For this calculation, extend the Gaussian noise model to all $t\geq1$
and allow any $\tau\geq1$.
With $q$ averaged responses per update and a fixed stepsize
$\eta\in(0,1]$, write $r=1-\eta\kappa$.
The limiting mean-square error is

\begin{equation}
\lim_{t\to\infty}\E x_t^2
=
\frac{\eta\sigma^2}{q\kappa(2-\eta\kappa)}
\frac{1+r\chi}{1-r\chi}.
\label{eq:stationary-correlation}
\end{equation}
Indeed, expanding the recursion gives
$\E x_t^2=(\eta^2\sigma^2/q)
\sum_{i,j=0}^{t-2}r^{i+j}\chi^{|i-j|}$;
summing the resulting geometric series yields
\eqref{eq:stationary-correlation}.

The first factor in \eqref{eq:stationary-correlation} is the limiting error under independent noise. Temporal correlation multiplies it by
\[
\frac{1+r\chi}{1-r\chi}
=
\frac{2\tau-\eta\kappa(\tau-1)}
     {2+\eta\kappa(\tau-1)}.
\]
When $\eta\kappa\tau$ is small, this multiplier is approximately $\tau$. In this regime, the tracker averages observations over a period longer than the noise remains strongly correlated. The upper restriction on $W$ in Theorem~\ref{thm:correlated-lower} ensures that
$\eta\kappa\tau\leq1/4$ for the balancing stepsize, placing the matching result in this regime.

For a fixed positive stepsize, the limiting error instead approaches
$\sigma^2/(q\kappa^2)$ as $\tau\to\infty$.
Thus, the error does not continue to grow proportionally to $\tau$ when correlations persist much longer than the tracker's averaging period. The general upper bound can overestimate the correlation cost in that setting.

\subsection{Adaptation using target validation}
\label{sec:meta}

The tuned stepsizes above depend on the target variation and noise parameters, which may be unknown. We run several copies of the tracking iteration, each with a different constant stepsize, and form a weighted average of their predictions. After predicting at round $t$, we receive an additional observation of $x_t^*$ whose conditional mean, given the learner's history, is $x_t^*$. We score each tracker by its squared distance to this observation and assign larger weights to trackers with smaller cumulative scores. 

\paragraph{Additional target observations.}
Let $\cX$ be a known nonempty closed convex set containing every
fixed point $x_t^*$ and contained in a ball of known radius $B>0$.
Let $\cG_t$ contain all information available before the validation
observation at round $t$, including the history, the internal iterates
of all trackers, and their current weights.

\begin{assumption}[Target validation]
\label{ass:validation}
At each round $t=1,\ldots,T$, after forming its prediction, the learner
receives a validation vector $Y_t\in\cX$ satisfying
\begin{equation}
\E[Y_t\mid\cG_t]=x_t^*.
\label{eq:validation}
\end{equation}
For $t<T$, conditional on $\cG_t$, $Y_t$ is independent of the
collection of operator samples used to update the trackers at round $t$.
\end{assumption}

\paragraph{Combining the trackers.}
Set $J=\lceil\log_2T\rceil$ and run $J+1$ trackers with stepsizes
$\eta_j=2^{-j}$, $j=0,\ldots,J$. We refer to these trackers as experts. Expert $j$ maintains an internal iterate $x_{t,j}$ and predicts
$\widetilde x_{t,j}=\Pi_{\cX}(x_{t,j})$, where $\Pi_{\cX}$ is the nearest-point projection onto $\cX$ with respect to the norm $\norm{\cdot}$
introduced in Section~\ref{sec:model}. The internal iterates follow the original affine update; projection is used to bound the predictions that are combined and scored.

At each update, expert $j$ receives
$\widehat F_{t,j}(x_{t,j})=F_t(x_{t,j})+\xi_{t+1,j}$.
For each fixed $j$, assume that its noise sequence satisfies
Assumption~\ref{ass:covariance} with the same parameters $(\sigma^2,\tau)$.
Different experts may share update samples, and their noise sequences need not be independent. Algorithm~\ref{alg:meta-track} combines the projected predictions by exponential weighting. Experts whose past predictions were closer to the validation observations receive larger weights.

\begin{algorithm}[H]
\caption{\textsc{Meta-Track}: combining trackers with target validation}
\label{alg:meta-track}
\begin{algorithmic}[1]
\STATE Set $J=\lceil\log_2T\rceil$ and $\lambda=1/(8B^2)$.
\STATE For each $j=0,\ldots,J$, set $\eta_j=2^{-j}$,
$p_{1,j}=1/(J+1)$, and $x_{1,j}=x_1$.
\FOR{$t=1,\ldots,T$}
\STATE Set $\widetilde x_{t,j}=\Pi_{\cX}(x_{t,j})$ for every $j$.
\STATE Predict $\bar x_t=\sum_{j=0}^{J}p_{t,j}\widetilde x_{t,j}$.
\STATE Observe the validation vector $Y_t$.
\IF{$t<T$}
\STATE Set
$p_{t+1,j}\propto p_{t,j}
\exp\{-\lambda\norm{\widetilde x_{t,j}-Y_t}^2\}$
and normalize the weights to sum to one.
\STATE Obtain the operator observations
$\widehat F_{t,j}(x_{t,j})$ for all experts.
\STATE Update
$x_{t+1,j}=(1-\eta_j)x_{t,j}
+\eta_j\widehat F_{t,j}(x_{t,j})$
for every $j$.
\ENDIF
\ENDFOR
\end{algorithmic}
\end{algorithm}

\begin{theorem}[Adaptation with target validation]
\label{thm:meta}
Suppose Assumptions~\ref{ass:contraction} and~\ref{ass:validation} hold,
and each expert's noise sequence satisfies
Assumption~\ref{ass:covariance} with the same parameters
$(\sigma^2,\tau)$.
Then Algorithm~\ref{alg:meta-track} satisfies
\begin{subequations}
\label{eq:meta-oracle}
\begin{align}
\sum_{t=1}^T\E\norm{\bar x_t-x_t^*}^2
&\leq
\min_{0\leq j\leq J}
\sum_{t=1}^T\E\norm{x_{t,j}-x_t^*}^2
+8B^2\log(J+1),
\label{eq:meta-grid}\\
&\leq
4\inf_{\eta\in[1/T,1]}
\left\{
\frac{\norm{x_1-x_1^*}^2}{\eta\kappa}
+\frac{\eta\tau\sigma^2T}{\kappa(2-\eta\kappa)}
+\frac{W_T}{\eta^2\kappa^2}
\right\}
+8B^2\log(J+1).
\label{eq:meta-bound}
\end{align}
\end{subequations}
If $x_1=x_1^*$, $W_T>0$, $\tau\sigma^2>0$, and the balancing stepsize
$\eta_{\rm bal}$ from Corollary~\ref{cor:tuned} lies in $[1/T,1]$, then
\[
\sum_{t=1}^T\E\norm{\bar x_t-x_t^*}^2
<
8\frac{(\tau\sigma^2T)^{2/3}W_T^{1/3}}{\kappa^{4/3}}
+8B^2\log(1+\lceil\log_2T\rceil).
\]
\end{theorem}

Equation~\ref{eq:meta-grid} bounds the combined predictor's cumulative
expected squared error by the smallest such error among the $J+1$
fixed-stepsize trackers, plus $8B^2\log(J+1)$.
Equation~\ref{eq:meta-bound} gives an explicit guarantee in terms of
initialization, noise, and target variation. It is obtained by applying
Theorem~\ref{thm:upper} to the trackers and comparing the stepsize grid
with all $\eta\in[1/T,1]$, which introduces the factor four.
Appendix~\ref{app:meta} gives the proof.

The algorithm uses $J+1=O(\log T)$ trackers and does not require
knowledge of $W_T$, $\sigma^2$, or $\tau$.
With separate pointwise queries, it uses $J+1$ operator calls per
update. It also requires one validation observation per prediction
round.
Excluding operator evaluations and projections, computing the tracker
updates, combining their predictions, and updating their weights
requires $O(T(J+1)d)$ arithmetic operations, assuming that squared
norms can be computed in $O(d)$ operations.

\paragraph{Direct smoothing of validation observations.}
The additional target observations also support a simpler tracker that uses no operator evaluations. For a fixed $a\in(0,1]$, initialize $u_1=x_1$ and update
\[
u_{t+1}=(1-a)u_t+aY_t,
\qquad t=1,\ldots,T-1.
\]
Conditional unbiasedness and boundedness imply
$\E[\norm{Y_t-x_t^*}^2\mid\cG_t]\leq B^2$.
Applying Theorem~\ref{thm:upper} to this recursion gives
\[
\sum_{t=1}^T\E\norm{u_t-x_t^*}^2
\leq
\frac{\norm{x_1-x_1^*}^2}{a}
+\frac{aB^2T}{2-a}
+\frac{W_T}{a^2}.
\]
Appendix~\ref{app:meta} gives the details.

At $a=\eta\kappa$, this bound has the same initialization and movement terms as the operator-tracking bound in Theorem~\ref{thm:upper}. The respective noise coefficients are $B^2$ for direct smoothing and $\tau\sigma^2/\kappa^2$ for operator tracking. Thus, the operator-tracking bound has a smaller noise term when $\tau\sigma^2/\kappa^2<B^2$. This comparison concerns the upper bounds; it does not establish that either method always has smaller actual error.

\subsection{Applications and extensions}
\label{sec:td}

Appendix~\ref{app:td} applies our results to synchronous tabular TD
when rewards and transition probabilities change. This application
requires a simulator that can sample from every state and assumes
that all transition kernels share the same invariant distribution.
When this distribution is known, we also show how additional
simulator rollouts provide the target observations needed for
stepsize adaptation and quantify their sampling cost.
Appendix~\ref{app:ttsa} extends the analysis to two coupled linear
updates, each with its own stepsize, and gives tracking bounds under
additional contraction assumptions.

\section{Discussion and comparison with prior work}
\label{sec:discussion}

Our analysis builds on the deterministic tracking guarantee of
\citet{hadiji2026tracking}. For the affine updates considered here,
Theorem~\ref{thm:upper} retains their initialization and target-movement
terms and adds a noise term that accounts for temporal correlation.
Earlier stochastic analyses already describe the tradeoff between
noise reduction and responsiveness to change, including in nonlinear
settings~\citep{bastianello2024stochastic,cutler2023distributional}.
The deterministic affine structure allows us to separate the effects
of initialization and target movement from accumulated observation
noise. We use this separation to obtain a bound in terms of squared
target variation and the covariance condition in
Assumption~\ref{ass:covariance}.
Table~\ref{tab:comparison} in Appendix~\ref{app:comparison} summarizes
the assumptions and guarantees of selected prior results.

The Gaussian benchmarks clarify when this bound is sharp.
For independent observations, the dependence on the horizon and
squared target variation agrees with classical results for estimating
sequences with bounded squared increments~\citep{sadhanala2016total}.
Our formulation makes the dependence on the contraction strength and
number of operator queries explicit. It also identifies variation
ranges in which a suitably chosen constant stepsize attains the
smallest possible worst-case error up to constant factors.
The correlated Gaussian benchmark further shows that the
$\tau^{2/3}$ dependence in the tuned bound is unavoidable in the
matching range described in Section~\ref{sec:correlated-lower}.
These optimality statements concern the specified scalar Gaussian
models and variation ranges.

The adaptive result addresses how to choose among stepsizes when
the noise level and target variation are unknown. Adaptation to
unknown variation is already possible in scalar forecasting:
\citet{baby2019online} obtain guarantees directly from noisy target
observations, without a separate validation stream. Our method
combines affine trackers using standard exponential
weighting~\citep{cesabianchi2006prediction} and requires additional
bounded, unbiased observations of the current fixed point.
Theorem~\ref{thm:meta} compares its cumulative expected error with
that of the best fixed-stepsize tracker in the collection.
The additional observations also support direct smoothing, as
discussed in Section~\ref{sec:meta}.

For policy evaluation, the sampling model is an essential distinction.
\citet{walton2020adiabatic} studies asynchronous TD and Q-learning
using observations from a changing Markov trajectory.
Our TD application assumes simulator access from every state and
transition kernels that share a common invariant distribution.
Under these conditions, we obtain cumulative value-estimation
bounds and quantify the additional sampling cost of adaptation.
Extending the analysis to observations from a single changing
trajectory, and adapting the stepsize using operator observations
alone, remain directions for further work.

\subsection*{AI use statement}
Generative AI tools were used to help draft portions of the text, identify related literature, and prepare \LaTeX. The authors remain responsible for independently checking every cited source, theorem statement, algebraic derivation, and proof, and for the final content of the manuscript. No experimental data were generated by AI.

\subsection*{Reproducibility statement}
Sections~\ref{sec:model}--\ref{sec:results} state the assumptions,
algorithms, and constants for every main-text result and specify
which observations are available when each prediction is formed. The appendix gives complete proofs and states the additional localization condition used for the nested fast--slow extension. The paper contains no experiments or datasets.

\bibliography{references}
\bibliographystyle{iclr2027_conference}

\appendix

\section*{Appendix roadmap}

We begin with the preliminary lemmas and the proof of the main
tracking bound, including the tuned-stepsize guarantee. We then
establish the Gaussian lower bounds for independent and temporally
correlated noise and identify the conditions under which they match
the upper bounds. Appendix~\ref{app:meta} contains the proof of the
adaptive tracking result and the analysis of direct smoothing using
validation observations. In Appendix~\ref{app:td}, we apply our
results to synchronous tabular TD, construct the unbiased validation
observations needed for adaptation, and quantify their sampling cost.
Appendix~\ref{app:ttsa} extends the analysis to linear recursions
with two stepsizes, considering both a stacked contraction and a
nested fast--slow system. Finally, Appendix~\ref{app:comparison}
compares our assumptions and guarantees with selected prior work.

\section{Preliminary lemmas}

We first collect the deterministic filter bounds, the covariance bound, and the exponential-weighting inequality used later.

\begin{lemma}[Stable products]
\label{lem:stable-products}
Let $K_1,\ldots,K_{T-1}$ be deterministic matrices, and satisfy $\norm{K_t}_{\op}\leq\rho<1$. For $t>s$, define $\Phi(t,s)=K_{t-1}\cdots K_s$, and let $\Phi(s,s)=I$. Then $\norm{\Phi(t,s)}_{\op}\leq\rho^{t-s}$ and
\[
\sum_{t=1}^T\norm{\Phi(t,1)u}^2
\leq\frac{\norm{u}^2}{1-\rho^2}.
\]
\end{lemma}

\begin{proof}
Submultiplicativity gives $\norm{\Phi(t,s)}_{\op}\leq\rho^{t-s}$. Hence
\[
\sum_{t=1}^T\norm{\Phi(t,1)u}^2
\leq\norm{u}^2\sum_{k=0}^{T-1}\rho^{2k}
\leq\frac{\norm{u}^2}{1-\rho^2}.
\]
\end{proof}

\begin{lemma}[Deterministic contraction tracking]
\label{lem:deterministic-tracking}
Suppose $d_{t+1}=K_td_t-u_t$ and $\norm{K_t}_{\op}\leq\rho<1$ for $t=1,\ldots,T-1$. Then
\[
\sum_{t=1}^T\norm{d_t}^2
\leq
\frac{\norm{d_1}^2}{1-\rho}
+\frac{1}{(1-\rho)^2}\sum_{t=1}^{T-1}\norm{u_t}^2.
\]
\end{lemma}

\begin{proof}
For nonnegative $a,b$,
\[
(\rho a+b)^2
\leq \rho a^2+\frac{b^2}{1-\rho},
\]
because the right-hand side minus the left-hand side equals
$\rho(\sqrt{1-\rho}\,a-b/\sqrt{1-\rho})^2$. Therefore
\[
\norm{d_{t+1}}^2
\leq
\rho\norm{d_t}^2+\frac{\norm{u_t}^2}{1-\rho}.
\]
Let $S_T=\sum_{t=1}^T\norm{d_t}^2$ and $U_T=\sum_{t=1}^{T-1}\norm{u_t}^2$. Summing the preceding inequality for $t=1,\ldots,T-1$ gives
\[
S_T-\norm{d_1}^2
\leq
\rho\bigl(S_T-\norm{d_T}^2\bigr)+\frac{U_T}{1-\rho}.
\]
Dropping the nonpositive term $-\rho\norm{d_T}^2$, rearranging, and dividing by $1-\rho$ prove the claim.
\end{proof}

\begin{lemma}[Stable convolution]
\label{lem:stable-convolution}
For vectors $d_1,\ldots,d_{T-1}$ and $\rho\in[0,1)$,
\[
\sum_{t=2}^T\left(\sum_{s=1}^{t-1}\rho^{t-1-s}\norm{d_s}\right)^2
\leq\frac{1}{(1-\rho)^2}\sum_{s=1}^{T-1}\norm{d_s}^2.
\]
Consequently, under the hypotheses of Lemma~\ref{lem:stable-products},
\[
\sum_{t=2}^T\norm{\sum_{s=1}^{t-1}\Phi(t,s+1)d_s}^2
\leq\frac{1}{(1-\rho)^2}\sum_{s=1}^{T-1}\norm{d_s}^2.
\]
\end{lemma}

\begin{proof}
Let $a_s=\norm{d_s}$. By the Cauchy--Schwarz inequality,
for every $t=2,\ldots,T$,
\begin{align*}
\left(\sum_{s=1}^{t-1}\rho^{t-1-s}a_s\right)^2
&\leq
\left(\sum_{s=1}^{t-1}\rho^{t-1-s}\right)
\left(\sum_{s=1}^{t-1}\rho^{t-1-s}a_s^2\right)\\
&\leq
\frac{1}{1-\rho}
\sum_{s=1}^{t-1}\rho^{t-1-s}a_s^2,
\end{align*}
where the last inequality follows from the geometric
series bound
$\sum_{s=1}^{t-1}\rho^{t-1-s}\leq(1-\rho)^{-1}$.

Summing over $t$ and exchanging the order of summation,
we obtain
\begin{align*}
&\sum_{t=2}^T
\left(\sum_{s=1}^{t-1}\rho^{t-1-s}a_s\right)^2\\
&\quad\leq
\frac{1}{1-\rho}
\sum_{t=2}^T\sum_{s=1}^{t-1}
\rho^{t-1-s}a_s^2\\
&\quad=
\frac{1}{1-\rho}
\sum_{s=1}^{T-1}a_s^2
\sum_{t=s+1}^T\rho^{t-1-s}\\
&\quad\leq
\frac{1}{(1-\rho)^2}
\sum_{s=1}^{T-1}a_s^2,
\end{align*}
which proves the first inequality.

For the matrix inequality, Lemma~\ref{lem:stable-products}
gives
$\norm{\Phi(t,s+1)}_{\op}\leq\rho^{t-1-s}$.
Hence, by the triangle inequality,
\begin{align*}
\norm{\sum_{s=1}^{t-1}\Phi(t,s+1)d_s}
&\leq
\sum_{s=1}^{t-1}
\norm{\Phi(t,s+1)d_s}\\
&\leq
\sum_{s=1}^{t-1}
\rho^{t-1-s}\norm{d_s}.
\end{align*}
Squaring both sides, summing over $t=2,\ldots,T$,
and applying the first inequality completes the proof.
\end{proof}

\begin{lemma}[Accumulated oracle error]
\label{lem:noise-accumulation}
Under Assumption~\ref{ass:covariance} and the hypotheses of Lemma~\ref{lem:stable-products},
\[
\sum_{t=2}^T\E\norm{\sum_{s=1}^{t-1}\Phi(t,s+1)\xi_{s+1}}^2
\leq\frac{\tau\sigma^2(T-1)}{1-\rho^2}.
\]
\end{lemma}

\begin{proof}
Fix $t$. Apply~\eqref{eq:integrated-covariance} with $C_s=\Phi(t,s+1)$ for $s<t$ and $C_s=0$ for $s\geq t$. Lemma~\ref{lem:stable-products} gives
\[
\E\norm{\sum_{s=1}^{t-1}\Phi(t,s+1)\xi_{s+1}}^2
\leq\tau\sigma^2\sum_{s=1}^{t-1}\rho^{2(t-1-s)}
\leq\frac{\tau\sigma^2}{1-\rho^2}.
\]
Finally, summing over $t=2,\ldots,T$ proves the result.
\end{proof}

\begin{proposition}[Geometric centered correlation implies integrated covariance]
\label{prop:geometric-correlation}
Suppose $\E[\xi_{t+1}]=0$ for every $t=1,\ldots,T-1$ and, for all deterministic linear maps $C_s,C_r:\R^d\to\R^d$,
\[
\abs{\E\ip{C_s\xi_{s+1}}{C_r\xi_{r+1}}}
\leq\sigma^2\chi_{\rm corr}^{|s-r|}\norm{C_s}_{\op}\norm{C_r}_{\op},
\qquad \chi_{\rm corr}\in[0,1).
\]
Then Assumption~\ref{ass:covariance} holds with
$\tau=(1+\chi_{\rm corr})/(1-\chi_{\rm corr})$.
\end{proposition}

\begin{proof}
We verify the two conditions in
Assumption~\ref{ass:covariance}.

First, taking $s=r$ and $C_s=C_r=I$ in the
assumed covariance bound gives
\begin{align*}
\E\norm{\xi_{s+1}}^2\leq\sigma^2.
\end{align*}
Together with the assumed mean-zero property,
this establishes the first condition.

Next, fix any $1\leq n\leq T-1$ and deterministic
matrices $C_1,\ldots,C_n$. Expanding the squared
norm and taking expectations yields
\begin{align*}
\E\norm{\sum_{s=1}^n C_s\xi_{s+1}}^2
&=\sum_{s=1}^n\sum_{r=1}^n
\E\ip{C_s\xi_{s+1}}{C_r\xi_{r+1}}\\
&\leq\sigma^2
\sum_{s=1}^n\sum_{r=1}^n
\chi_{\rm corr}^{|s-r|}
\norm{C_s}_{\op}\norm{C_r}_{\op},
\end{align*}
where the inequality follows from the assumed
covariance bound.

Using $2ab\leq a^2+b^2$, we have
\begin{align*}
&\sum_{s=1}^n\sum_{r=1}^n
\chi_{\rm corr}^{|s-r|}
\norm{C_s}_{\op}\norm{C_r}_{\op}\\
&\quad\leq
\frac12\sum_{s=1}^n\sum_{r=1}^n
\chi_{\rm corr}^{|s-r|}
\left(\norm{C_s}_{\op}^2+\norm{C_r}_{\op}^2\right)\\
&\quad=
\sum_{s=1}^n\norm{C_s}_{\op}^2
\sum_{r=1}^n\chi_{\rm corr}^{|s-r|}.
\end{align*}
The equality follows by symmetry in $s$ and $r$.

For each fixed $s$, the inner sum satisfies
\begin{align*}
\sum_{r=1}^n\chi_{\rm corr}^{|s-r|}
&=1+\sum_{k=1}^{s-1}\chi_{\rm corr}^k
+\sum_{k=1}^{n-s}\chi_{\rm corr}^k\\
&\leq 1+2\sum_{k=1}^{\infty}\chi_{\rm corr}^k\\
&=1+\frac{2\chi_{\rm corr}}{1-\chi_{\rm corr}}\\
&=\frac{1+\chi_{\rm corr}}{1-\chi_{\rm corr}}.
\end{align*}
Combining these inequalities gives
\begin{align*}
\E\norm{\sum_{s=1}^n C_s\xi_{s+1}}^2
&\leq
\sigma^2\frac{1+\chi_{\rm corr}}{1-\chi_{\rm corr}}
\sum_{s=1}^n\norm{C_s}_{\op}^2\\
&=\tau\sigma^2
\sum_{s=1}^n\norm{C_s}_{\op}^2,
\end{align*}
where
$\tau=(1+\chi_{\rm corr})/(1-\chi_{\rm corr})$.
This establishes~\eqref{eq:integrated-covariance}
and completes the proof.
\end{proof}

\begin{lemma}[Exp-concavity of squared loss]
\label{lem:exp-concavity}
Let $B>0$, let $\cX$ be convex and contained in a ball of radius $B$, and let $y\in\cX$. For every $0<\lambda\leq1/(8B^2)$, the map
$x\mapsto\exp(-\lambda\norm{x-y}^2)$ is concave on $\cX$. Consequently, for $x_j\in\cX$ and a probability vector $p$,
\[
\norm{\sum_jp_jx_j-y}^2
\leq-\lambda^{-1}
\log\sum_jp_j\exp(-\lambda\norm{x_j-y}^2).
\]
\end{lemma}

\begin{proof}
We first show that the function
\begin{align*}
f(x)=\exp(-\lambda\norm{x-y}^2)
\end{align*}
is concave on $\cX$ whenever
$0<\lambda\leq 1/(8B^2)$.

Fix any $x,x'\in\cX$ and consider the line segment
\begin{align*}
z(t)=(1-t)x+tx',
\qquad t\in[0,1].
\end{align*}
Since $\cX$ is convex, $z(t)\in\cX$ for every
$t\in[0,1]$. Moreover, because $\cX$ is contained
in a ball of radius $B$ and $y\in\cX$,
\begin{align*}
\norm{z(t)-y}\leq 2B.
\end{align*}

Let $h=x'-x$ and define
\begin{align*}
g(t)=f(z(t))
=\exp(-\lambda\norm{z(t)-y}^2).
\end{align*}
Since the norm is induced by an inner product
and $z'(t)=h$, differentiation gives
\begin{align*}
g'(t)
&=-2\lambda g(t)\ip{z(t)-y}{h},\\
g''(t)
&=g(t)\left(
4\lambda^2\ip{z(t)-y}{h}^2
-2\lambda\norm{h}^2
\right).
\end{align*}
By the Cauchy--Schwarz inequality,
\begin{align*}
\ip{z(t)-y}{h}^2
&\leq\norm{z(t)-y}^2\norm{h}^2\\
&\leq 4B^2\norm{h}^2.
\end{align*}
Substituting this bound into the expression
for $g''(t)$ yields
\begin{align*}
g''(t)
&\leq g(t)
\left(16\lambda^2B^2-2\lambda\right)
\norm{h}^2\\
&=2\lambda g(t)
\left(8\lambda B^2-1\right)
\norm{h}^2\\
&\leq 0,
\end{align*}
where the last inequality follows from
$\lambda\leq 1/(8B^2)$.

Thus, $g$ is concave on $[0,1]$.
Since $x$ and $x'$ were arbitrary, $f$ is
concave on $\cX$.

Now let $x_j\in\cX$ and let $(p_j)$ be a
probability vector. By convexity of $\cX$,
the weighted average $\bar x=\sum_jp_jx_j$
belongs to $\cX$. Applying Jensen's
inequality to the concave function $f$ gives
\begin{align*}
\exp(-\lambda\norm{\bar x-y}^2)
&\geq\sum_jp_j
\exp(-\lambda\norm{x_j-y}^2).
\end{align*}
Taking logarithms of both sides and
dividing by $-\lambda<0$ reverses the
inequality, yielding
\begin{align*}
\norm{\sum_jp_jx_j-y}^2
&\leq-\frac{1}{\lambda}
\log\sum_jp_j
\exp(-\lambda\norm{x_j-y}^2).
\end{align*}
This completes the proof.
\end{proof}

\section{Proof of the affine tracking upper bound}

We combine the deterministic tracking lemma with the covariance bound for the propagated noise.

\begin{proof}[Proof of Theorem~\ref{thm:upper}]
Let $e_t=x_t-x_t^*$ and $\Delta_t=x_{t+1}^*-x_t^*$. Since $F_t(x_t^*)=x_t^*$,
\[
\begin{aligned}
e_{t+1}
&=(1-\eta)(x_t-x_t^*)+\eta\bigl(F_t(x_t)-F_t(x_t^*)\bigr)
+\eta\xi_{t+1}-\Delta_t\\
&=K_te_t+\eta\xi_{t+1}-\Delta_t,
\end{aligned}
\]
where $K_t=(1-\eta)I+\eta M_t$. Here we used the
affine identity $F_t(x_t)-F_t(x_t^*)
=M_t(x_t-x_t^*)$.

Assumption~\ref{ass:contraction} gives
$\norm{K_t}_{\op}\leq(1-\eta)+\eta(1-\kappa)=1-\eta\kappa=: \rho<1$.

Split $e_t=d_t+n_t$, where
\[
d_1=e_1,\quad n_1=0,\qquad
d_{t+1}=K_td_t-\Delta_t,\qquad
n_{t+1}=K_tn_t+\eta\xi_{t+1}.
\]
All $K_t$, $d_t$, and $\Delta_t$ are deterministic. Marginal centering in Assumption~\ref{ass:covariance} gives $\E[n_t]=0$, and therefore
\[
\E\norm{e_t}^2=\norm{d_t}^2+\E\norm{n_t}^2.
\]
Lemma~\ref{lem:deterministic-tracking} yields
\[
\sum_{t=1}^T\norm{d_t}^2
\leq
\frac{\norm{e_1}^2}{1-\rho}
+\frac{W_T}{(1-\rho)^2}.
\]
Moreover,
\[
n_t=\eta\sum_{s=1}^{t-1}\Phi(t,s+1)\xi_{s+1}.
\]
Lemma~\ref{lem:noise-accumulation} gives
\[
\sum_{t=1}^T\E\norm{n_t}^2
\leq\frac{\eta^2\tau\sigma^2(T-1)}{1-\rho^2}
\leq\frac{\eta^2\tau\sigma^2T}{1-\rho^2}.
\]
Combining the deterministic and stochastic bounds and using
$1-\rho=\eta\kappa$ and
$1-\rho^2=\eta\kappa(2-\eta\kappa)$ proves~\eqref{eq:upper-main}.
\end{proof}

\begin{proof}[Proof of Corollary~\ref{cor:tuned}]
When $x_1=x_1^*$, Theorem~\ref{thm:upper} and $2-\eta\kappa\geq1$ give
\[
\cE_T\leq a\eta+b\eta^{-2},
\qquad
a=\frac{\tau\sigma^2T}{\kappa},\quad b=\frac{W_T}{\kappa^2}.
\]
The unconstrained minimizer of this simpler bound is
$(2b/a)^{1/3}=\eta_{\rm bal}$. Substituting it gives
\[
a\eta_{\rm bal}+b\eta_{\rm bal}^{-2}
=(2^{1/3}+2^{-2/3})a^{2/3}b^{1/3},
\]
which is the claimed bound.
\end{proof}

\section{Proof of the Gaussian lower bound}

\paragraph{Proof of Theorem~\ref{thm:lower}.}
We prove the lower bound by reducing the oracle model to a
Gaussian sequence estimation problem. We first show that the
query locations provide no additional information about the
target and that the observations at each round can be replaced
by their average. We then construct a family of triangular
target sequences that satisfy $\theta_1=0$ and the variation
budget but are difficult to distinguish from noisy observations.
A testing argument converts this difficulty into the desired
lower bound on the squared estimation error.

\subsection{Reduction to a Gaussian sequence experiment}
\label{app:gaussian-reduction}
Let $\mathsf O_{t,i}(x)$ denote the $i$th response to a query $x$ at round $t$. A possibly adaptive query $x_{t,i}$ produces
\[
Z_{t,i}:=\frac{\mathsf O_{t,i}(x_{t,i})-(1-\kappa)x_{t,i}}{\kappa}
=\theta_t+\frac{\varepsilon_{t,i}}{\kappa}.
\]
The transformation removes the known contribution of the query
location, leaving a Gaussian observation of $\theta_t$ whose
distribution does not depend on $x_{t,i}$. Since $\kappa$ and
$x_{t,i}$ are known to the learner, the transformation is
invertible: the original response can be recovered from
$Z_{t,i}$. This remains true for adaptive queries, since the
learner can reconstruct its past responses and choose the same
subsequent query locations. Thus, working with the transformed
observations does not lose information about the target. For
the lower bound, we may also give the learner exactly $q$
observations at every round, since additional observations
can only improve estimation accuracy.

\begin{lemma}[Sufficiency of the per-round sample mean]
For each $t=1,\ldots,T$, suppose the learner observes
$Z_{t,1},\ldots,Z_{t,q}$, where $Z_{t,i}\sim\mathcal N(\theta_t,\sigma^2/\kappa^2)$, independently across queries and rounds. Define
\begin{align*}
\bar Z_t
&=\frac{1}{q}\sum_{i=1}^q Z_{t,i}
\sim\mathcal N(\theta_t,\nu^2),
\qquad
\nu^2=\frac{\sigma^2}{q\kappa^2}.
\end{align*}
For estimating $\theta_{1:T}$ under squared loss, it is
sufficient to observe the sample means
$\bar Z_1,\ldots,\bar Z_T$. In particular, any estimator
based on all $qT$ observations can be replaced by an
estimator based only on these sample means, without
increasing its expected squared error.
\end{lemma}

\begin{proof}
We show that the observations contain no additional
information about $\theta_{1:T}$ beyond their sample
means.

Fix a round $t$ and write $v^2=\sigma^2/\kappa^2$.
For any possible observations $z_{t,1},\ldots,z_{t,q}$,
their joint density is proportional to
\begin{align*}
\exp\left(
-\frac{1}{2v^2}
\sum_{i=1}^q(z_{t,i}-\theta_t)^2
\right).
\end{align*}
The sum in the exponent can be written as
\begin{align*}
\sum_{i=1}^q(z_{t,i}-\theta_t)^2
&=\sum_{i=1}^q(z_{t,i}-\bar z_t)^2
+q(\bar z_t-\theta_t)^2,
\end{align*}
where $\bar z_t=q^{-1}\sum_{i=1}^q z_{t,i}$. To understand why the sample mean is sufficient,
consider the two terms in the preceding identity.
The first describes how the observations vary around
their sample mean and does not depend on $\theta_t$.
The second contains all the dependence on $\theta_t$,
but involves the observations only through $\bar z_t$.

Now suppose we condition on the sample mean taking
a particular value $\bar z_t$. Among all possible
sets of observations with this average, the second
term is constant. It therefore cancels when computing
the conditional distribution of the observations.
The remaining term depends only on the deviations
from the sample mean and not on $\theta_t$.

Thus, once $\bar Z_t$ is known, the conditional
distribution of the original $q$ observations does
not depend on $\theta_t$. Since observations are
independent across rounds, the same conclusion holds
for the entire sequence: the conditional distribution
of all $qT$ observations given $\bar Z_{1:T}$ does
not depend on $\theta_{1:T}$.

We now show that using only the sample means does
not increase the estimation error. Let
$\widehat\theta$ be any estimator that uses all
$qT$ observations. We construct a new estimator
by taking the conditional expectation of
$\widehat\theta$ given the sample means:
\begin{align*}
\widetilde\theta
=\E_\theta[\widehat\theta\mid\bar Z_{1:T}].
\end{align*}
In other words, for a given collection of sample
means, $\widetilde\theta$ averages the estimates
produced by $\widehat\theta$ over the possible
values of the individual observations consistent
with those means.

Although the conditional expectation is written
under the distribution indexed by $\theta$, the
resulting estimator does not require knowledge
of $\theta$. Indeed, we have already shown that
the conditional distribution of the individual
observations given their sample means is
independent of $\theta$. Therefore, the
conditional averaging above can be performed
using only $\bar Z_{1:T}$, and the same estimator
$\widetilde\theta$ works for every target sequence.

It remains to compare the estimation errors.
Fix any target sequence $\theta$. Since $\theta$
is deterministic, we have
\begin{align*}
\widetilde\theta-\theta
&=\E_\theta[
\widehat\theta-\theta
\mid\bar Z_{1:T}].
\end{align*}
The squared Euclidean norm is convex, so
Jensen's inequality gives
\begin{align*}
\norm{\widetilde\theta-\theta}_2^2
&=
\norm{
\E_\theta[\widehat\theta-\theta
\mid\bar Z_{1:T}]
}_2^2\\
&\leq
\E_\theta[
\norm{\widehat\theta-\theta}_2^2
\mid\bar Z_{1:T}].
\end{align*}
Taking expectations on both sides and using
the law of total expectation yields
\begin{align*}
\E_\theta\norm{\widetilde\theta-\theta}_2^2
&\leq
\E_\theta\left[
\E_\theta[
\norm{\widehat\theta-\theta}_2^2
\mid\bar Z_{1:T}]
\right]\\
&=
\E_\theta\norm{\widehat\theta-\theta}_2^2.
\end{align*}

Thus, for every estimator based on the full
observations, we have constructed an estimator
based only on the sample means whose expected
squared error is no larger, for every
$\theta$. This proves that restricting attention
to the sample means does not increase the
minimax estimation risk.
\end{proof}

It remains to lower bound the offline Gaussian sequence model
$\bar Z_t\sim\mathcal N(\theta_t,\nu^2)$.

\subsection{Triangular packing}
\label{app:triangular-packing}
\begin{figure}[t]
\centering
\begin{tikzpicture}[scale=1]
  \draw[->] (0,0) -- (6.2,0) node[right] {$i$};
  \draw[->] (0,-0.2) -- (0,2.2);
  \draw[dashed] (0.8,-0.15) -- (0.8,1.9);
  \draw[dashed] (5.2,-0.15) -- (5.2,1.9);
  \draw[thick] (0.8,0) -- (3,1.6) -- (5.2,0);
  \node[below] at (0.8,0) {$1$};
  \node[below] at (3,0) {$r+1$};
  \node[below] at (5.2,0) {$m$};
  \node[left] at (0,1.6) {$1$};
  \node[below] at (3,-0.45) {support of $\psi$};
  \node[above] at (3.2,1.65) {$\psi_i$};
\end{tikzpicture}
\caption{The triangular sequence $\psi$ on a single block of length $m=2r+1$. It starts at $0$, rises linearly to $1$, and then decreases linearly back to $0$.}
\label{fig:triangle-psi}
\end{figure}
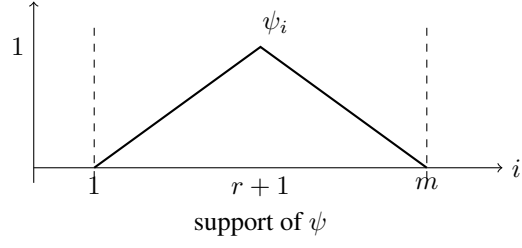

Let $m=2r+1$ be odd with $r\geq2$, and define $\psi\in\R^m$ by
\[
\psi_i=\frac{\min\{i-1,m-i\}}{r},
\qquad i=1,\ldots,m.
\]
Thus $\psi_1=\psi_m=0$ and the center equals one. Figure \ref{fig:triangle-psi} shows this function.

\begin{lemma}[Triangle geometry]
\label{lem:triangle-geometry}
For $m=2r+1$ with $r\geq2$,
\[
\frac{m}{6}\leq\norm{\psi}_2^2\leq m,
\qquad
\sum_{i=1}^{m-1}(\psi_{i+1}-\psi_i)^2=\frac{2}{r}\leq\frac{8}{m}.
\]
\end{lemma}

\begin{proof}
The nonzero values are
$1/r,2/r,\ldots,(r-1)/r,1,(r-1)/r,\ldots,1/r$, so
\[
\norm{\psi}_2^2
=1+\frac{2}{r^2}\sum_{j=1}^{r-1}j^2
=1+\frac{(r-1)(2r-1)}{3r}.
\]
For $r\geq2$, this lies between $(2r+1)/6=m/6$ and $m$. Each of the $2r$ nonzero increments has magnitude $1/r$, so $\sum_{i=1}^{m-1}(\psi_{i+1}-\psi_i)^2=\frac{2}{r}$. Finally, $m=2r+1\leq4r$ implies $2/r\leq8/m$.
\end{proof}

We now use the triangular sequence $\psi$ to construct
a family of target sequences $\theta^u$.
The goal is to obtain sequences with small squared variation
that are difficult to distinguish from noisy observations.

Fix an odd block length $m\leq T$ as above, and let
$K=\lfloor T/m\rfloor$. Divide the first $Km$ time
indices into $K$ consecutive blocks of length $m$:
\begin{align*}
\mathcal I_j
=\{(j-1)m+1,\ldots,jm\},
\qquad j=1,\ldots,K.
\end{align*}
For each block $j$, define a vector
$\psi^{(j)}\in\R^T$ by
\begin{align*}
\psi_t^{(j)}
=
\begin{cases}
\psi_{t-(j-1)m}, & t\in\mathcal I_j,\\
0, & t\notin\mathcal I_j.
\end{cases}
\end{align*}
Thus, $\psi^{(j)}$ agrees with $\psi$ on block
$j$ and is zero outside that block.

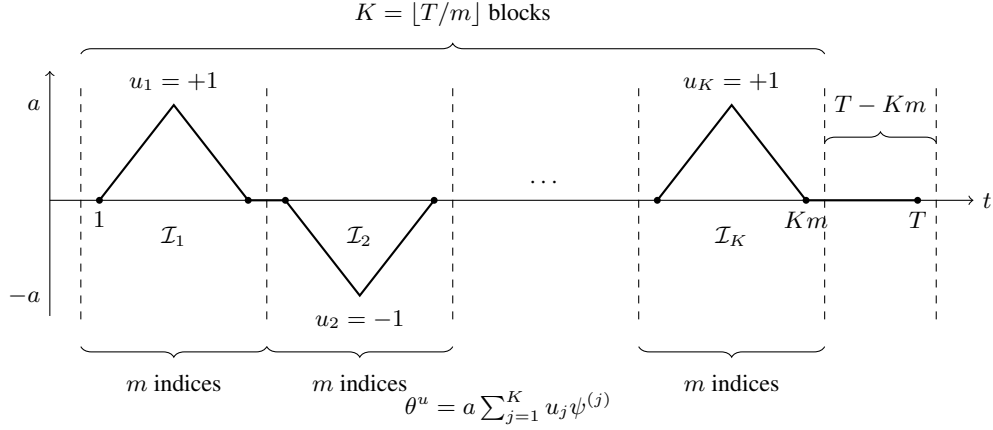
\begin{figure}[t]
\centering
\begin{tikzpicture}[x=0.82cm,y=0.90cm,font=\small]
  \draw[->] (0,0) -- (14.9,0) node[right] {$t$};
  \draw[->] (0,-1.7) -- (0,1.9);
  \foreach \x in {0.5,3.5,6.5,9.5,12.5,14.3}
    \draw[dashed,thin] (\x,-1.75) -- (\x,1.75);

  \draw[thick] (0.8,0) -- (2.0,1.4) -- (3.2,0)
               -- (3.8,0) -- (5.0,-1.4) -- (6.2,0);
  \node[above] at (8.0,0.05) {$\cdots$};
  \draw[thick] (9.8,0) -- (11.0,1.4) -- (12.2,0)
               -- (14.0,0);
  \foreach \x in {0.8,3.2,3.8,6.2,9.8,12.2,14.0}
    \fill (\x,0) circle (1.3pt);

  \node[below] at (2.0,-0.25) {$\mathcal I_1$};
  \node[below] at (5.0,-0.25) {$\mathcal I_2$};
  \node[below] at (11.0,-0.25) {$\mathcal I_K$};
  \node[above] at (2.0,1.48) {$u_1=+1$};
  \node[below] at (5.0,-1.48) {$u_2=-1$};
  \node[above] at (11.0,1.48) {$u_K=+1$};
  \node[left] at (0,1.4) {$a$};
  \node[left] at (0,-1.4) {$-a$};
  \node[below] at (0.8,-0.03) {$1$};
  \node[below] at (12.2,-0.03) {$Km$};
  \node[below] at (14.0,-0.03) {$T$};

  \draw[decorate,decoration={brace,amplitude=4pt}]
    (0.5,2.15) -- (12.5,2.15)
    node[midway,above=7pt] {$K=\lfloor T/m\rfloor\text{ blocks}$};
  \foreach \a/\b in {0.5/3.5,3.5/6.5,9.5/12.5}
    \draw[decorate,decoration={brace,mirror,amplitude=4pt}]
      (\a,-2.15) -- (\b,-2.15)
      node[midway,below=7pt] {$m\text{ indices}$};
  \draw[decorate,decoration={brace,amplitude=4pt}]
    (12.5,0.85) -- (14.3,0.85)
    node[midway,above=7pt] {$T-Km$};
  \node at (7.4,-3.05)
    {$\theta^u=a\sum_{j=1}^K u_j\psi^{(j)}$};
\end{tikzpicture}
\caption{Schematic of the signed triangular construction.
The first $Km$ indices form $K=\lfloor T/m\rfloor$ disjoint
blocks, each containing $m$ indices. Dashed lines mark the
boundaries between blocks; the zero-valued endpoints of
adjacent blocks are distinct coordinates. The sequence is
zero at the remaining $T-Km$ indices, where $0\leq T-Km<m$.
The figure illustrates a nonzero remainder; this final part
is absent when $Km=T$.}
\label{fig:theta-construction}
\end{figure}

For each sign vector $u=(u_1,\ldots,u_K)
\in\{-1,+1\}^K$, define
\begin{align*}
\theta^u
=a\sum_{j=1}^K u_j\psi^{(j)},
\qquad
a^2=\frac{\nu^2}{32m},
\end{align*}
where $a>0$. On block $j$, the resulting
sequence is $a\psi$ if $u_j=+1$ and
$-a\psi$ if $u_j=-1$. Any remaining
coordinates after time $Km$ are set to zero. Since $\psi_1=\psi_m=0$, each block begins
and ends at zero. In particular, every sequence
in this family satisfies $\theta_1^u=0$.
There are $2^K$ possible sequences, corresponding
to the different choices of signs. Figure \ref{fig:theta-construction} shows this construction.

For each $u$, let $\Pp_u$ denote the joint
distribution of the Gaussian observations
$\bar Z_1,\ldots,\bar Z_T$ when the true
target sequence is $\theta^u$.

\begin{lemma}[Budget and neighboring information]
\label{lem:packing-properties}
For every $u\in\{-1,+1\}^K$, the constructed
target sequence satisfies
\begin{align*}
\sum_{t=1}^{T-1}
(\theta_{t+1}^u-\theta_t^u)^2
\leq\frac{\nu^2T}{4m^3}.
\end{align*}
Moreover, if two sign vectors $u$ and $u'$
differ in exactly one coordinate, then
\begin{align*}
D_{\rm KL}(\Pp_u\Vert\Pp_{u'})
\leq\frac{1}{16}.
\end{align*}
\end{lemma}

\begin{proof}
We first bound the squared variation of
$\theta^u$. On block $j$, consecutive values
differ by
\begin{align*}
\theta_{(j-1)m+i+1}^u
-\theta_{(j-1)m+i}^u
=a u_j(\psi_{i+1}-\psi_i),
\end{align*}
for $i=1,\ldots,m-1$.

Since $u_j^2=1$, the squared increments
within each block sum to
\begin{align*}
a^2\sum_{i=1}^{m-1}
(\psi_{i+1}-\psi_i)^2.
\end{align*}
There is no additional variation between
blocks, because each block ends at zero
and the next block also begins at zero.
The same holds for any unused coordinates
after the last block.

Therefore, summing over all $K$ blocks gives
\begin{align*}
&\sum_{t=1}^{T-1}
(\theta_{t+1}^u-\theta_t^u)^2\\
&\quad=
Ka^2\sum_{i=1}^{m-1}
(\psi_{i+1}-\psi_i)^2\\
&\quad\leq
Ka^2\frac{8}{m}\\
&\quad\leq
\frac{T}{m}\frac{\nu^2}{32m}
\frac{8}{m}\\
&\quad=
\frac{\nu^2T}{4m^3},
\end{align*}
where the first inequality follows from
Lemma~\ref{lem:triangle-geometry}, and the
second uses $K\leq T/m$ and the definition
of $a^2$. This proves the variation bound.

We next bound the KL divergence. Suppose
$u$ and $u'$ differ only in coordinate $j$.
The corresponding target sequences are
identical outside block $j$. On that block,
one sequence equals $a\psi$ and the other
equals $-a\psi$. Consequently,
\begin{align*}
\theta^u-\theta^{u'}
=\pm 2a\psi^{(j)},
\end{align*}
and hence
\begin{align*}
\norm{\theta^u-\theta^{u'}}_2^2
=4a^2\norm{\psi}_2^2.
\end{align*}

Under $\Pp_u$, the observations are independent
Gaussian random variables with means
$\theta_t^u$ and common variance $\nu^2$.
The same variance applies under $\Pp_{u'}$,
but with means $\theta_t^{u'}$.
For two Gaussian distributions with the same
covariance matrix $\nu^2 I_T$, the KL divergence
equals the squared distance between their
mean vectors divided by $2\nu^2$. Therefore,
\begin{align*}
D_{\rm KL}(\Pp_u\Vert\Pp_{u'})
&=\frac{\norm{\theta^u-\theta^{u'}}_2^2}
{2\nu^2}\\
&=\frac{4a^2\norm{\psi}_2^2}{2\nu^2}\\
&=\frac{2a^2\norm{\psi}_2^2}{\nu^2}\\
&\leq\frac{2a^2m}{\nu^2}\\
&=\frac{1}{16},
\end{align*}
where we used $\norm{\psi}_2^2\leq m$ from
Lemma~\ref{lem:triangle-geometry} and
$a^2=\nu^2/(32m)$.

Thus, changing the sign of any single
block produces two target sequences whose
observation distributions have KL divergence
at most $1/16$, as claimed.
\end{proof}

\subsection{Testing reduction and completion}
\label{app:testing-reduction}
We now use the family of target sequences constructed above
to establish the lower bound. The main idea is that an
accurate estimate of $\theta^u$ should also identify the
sign of the triangular signal on each block. However,
Lemma~\ref{lem:packing-properties} shows that changing
the sign of one block produces two Gaussian observation
distributions that are difficult to distinguish. We use
this observation to obtain a lower bound on the
estimation error.

Let $\widehat\theta\in\R^T$ be any estimator of the
target sequence. We associate with this estimator
a rule for predicting the sign $u_j$ on each block.
Recall that the true target on block $\mathcal I_j$
is $a u_j\psi$, so
\begin{align*}
\ip{\theta^u_{\mathcal I_j}}{\psi}
=a u_j\norm{\psi}_2^2,
\end{align*}
where $\theta_{\mathcal I_j}^u$ denotes the
restriction of $\theta^u$ to block $\mathcal I_j$.
The sign of this inner product is therefore $u_j$.
Motivated by this observation, we define
\begin{align*}
\widehat u_j
=
\begin{cases}
+1, & \ip{\widehat\theta_{\mathcal I_j}}{\psi}\geq0,\\
-1, & \ip{\widehat\theta_{\mathcal I_j}}{\psi}<0.
\end{cases}
\end{align*}
We do not require this to be the optimal estimator
of $u_j$. We use it only to relate errors in
estimating the signs to the squared estimation
error of $\widehat\theta$.

Suppose that $\widehat u_j\neq u_j$. By the definition
of $\widehat u_j$, this implies
$u_j\ip{\widehat\theta_{\mathcal I_j}}{\psi}\leq0$.
Expanding the squared estimation error on this block,
we obtain
\begin{align*}
\norm{\widehat\theta_{\mathcal I_j}-a u_j\psi}_2^2
&=\norm{\widehat\theta_{\mathcal I_j}}_2^2
+a^2\norm{\psi}_2^2
-2a u_j\ip{\widehat\theta_{\mathcal I_j}}{\psi}\\
&\geq a^2\norm{\psi}_2^2.
\end{align*}
Thus, whenever the estimated sign is incorrect,
the squared estimation error on that block is at
least $a^2\norm{\psi}_2^2$.

Since the blocks are disjoint, we can sum this
inequality over all blocks. The squared errors
outside these blocks are nonnegative, so
\begin{align}
\norm{\widehat\theta-\theta^u}_2^2
&\geq
\sum_{j=1}^K
\norm{\widehat\theta_{\mathcal I_j}
-a u_j\psi}_2^2\nonumber\\
&\geq
a^2\norm{\psi}_2^2
\sum_{j=1}^K
\ind\{\widehat u_j\neq u_j\}.\label{eq:theta_lower_indic}
\end{align}
Therefore, a lower bound on the probability of
incorrectly identifying the signs will also give
a lower bound on the squared estimation error.

We next show that each sign has a non-negligible
probability of being estimated incorrectly.
Fix a block $j$ and fix the signs on all the
other blocks. Let $u^+$ and $u^-$ denote the two
sign vectors that agree on every coordinate except
$j$, where $u_j^+=+1$ and $u_j^-=-1$.

Consider the problem of distinguishing these two
possibilities from the Gaussian observations.
Let $A_j$ be the event that $\widehat u_j=+1$.
The average probability of making an incorrect
decision under the two possibilities is
\begin{align*}
&\frac12\left(
\Pp_{u^+}(A_j^c)+\Pp_{u^-}(A_j)
\right)\\
&\quad=
\frac12\left(
1-\Pp_{u^+}(A_j)+\Pp_{u^-}(A_j)
\right)\\
&\quad\geq
\frac12\left(
1-\operatorname{TV}(\Pp_{u^+},\Pp_{u^-})
\right),
\end{align*}
where $\operatorname{TV}$ denotes total-variation
distance. The last inequality follows because
total-variation distance bounds the difference
between the probabilities assigned to any event
by two distributions.

Since $u^+$ and $u^-$ differ in exactly one
coordinate, Lemma~\ref{lem:packing-properties}
gives
\begin{align*}
D_{\rm KL}(\Pp_{u^+}\Vert\Pp_{u^-})
\leq\frac1{16}.
\end{align*}
Applying Pinsker's inequality
\citep{tsybakov2009introduction}, we obtain
\begin{align*}
\operatorname{TV}(\Pp_{u^+},\Pp_{u^-})
&\leq
\sqrt{\frac12
D_{\rm KL}(\Pp_{u^+}\Vert\Pp_{u^-})}\\
&\leq\sqrt{\frac1{32}}
<\frac14.
\end{align*}
Consequently,
\begin{align*}
\frac12\left(
\Pp_{u^+}(\widehat u_j=-1)
+\Pp_{u^-}(\widehat u_j=+1)
\right)
\geq\frac14.
\end{align*}

This bound holds for every choice of the signs
on the other $K-1$ blocks. We can therefore
average over all such choices. If all $2^K$
sign vectors are assigned equal probability,
then, for every block $j$,
\begin{align*}
\frac1{2^K}
\sum_{u\in\{-1,+1\}^K}
\Pp_u(\widehat u_j\neq u_j)
\geq\frac14.
\end{align*}
In other words, even when all the other signs
are fixed, the observations do not allow the
learner to identify the remaining sign with
arbitrarily high accuracy.

We now return to the squared estimation error. Recall that the squared estimation error is at least
$a^2\norm{\psi}_2^2$ for each block whose sign is
estimated incorrectly. Taking expectations on both sides of \eqref{eq:theta_lower_indic} under
$\Pp_u$, using the fact that the expectation of an
indicator equals the probability of its event, and
then averaging over all $2^K$ sign vectors gives
\begin{align*}
&\frac1{2^K}
\sum_{u\in\{-1,+1\}^K}
\E_u\norm{\widehat\theta-\theta^u}_2^2\\
&\quad\geq
a^2\norm{\psi}_2^2
\sum_{j=1}^K
\frac1{2^K}
\sum_{u\in\{-1,+1\}^K}
\Pp_u(\widehat u_j\neq u_j)\\
&\quad\geq
\frac K4 a^2\norm{\psi}_2^2\\
&\quad\geq
\frac K4\frac{\nu^2}{32m}\frac m6\\
&\quad=
\frac{K\nu^2}{768}.
\end{align*}
Here, the second inequality uses the lower
bound on the sign-error probability, and the
third follows from
$\norm{\psi}_2^2\geq m/6$ in
Lemma~\ref{lem:triangle-geometry}, together
with $a^2=\nu^2/(32m)$.

It remains to choose the block length $m$ so
that every constructed target sequence belongs
to $\Theta_T(W)$. Define
\begin{align*}
m_0=\left(\frac{\nu^2T}{4W}\right)^{1/3},
\end{align*}
and let $m$ be the smallest odd integer satisfying
\begin{align*}
m\geq\max\{5,m_0\}.
\end{align*}
Under the variation range in
Theorem~\ref{thm:lower}, we have
\begin{align*}
\frac{16\nu^2}{T^2}\leq W
\leq\frac{\nu^2T}{32},
\end{align*}
which implies
\begin{align*}
2\leq m_0\leq\frac T4.
\end{align*}

We now verify two useful bounds on $m$.
If $m_0\leq5$, then $m=5$, so
$m\leq3m_0$ and $m\leq T/2$, since
$T\geq16$. If $m_0>5$, the definition
of $m$ gives $m\leq m_0+2$. Hence,
\begin{align*}
m&\leq3m_0,\\
m&\leq\frac T4+2\leq\frac T2.
\end{align*}
Thus, in both cases,
\begin{align*}
m_0\leq m\leq3m_0,
\qquad m\leq\frac T2.
\end{align*}

By Lemma~\ref{lem:packing-properties} and
the fact that $m\geq m_0$, every constructed
sequence satisfies
\begin{align*}
\sum_{t=1}^{T-1}
(\theta_{t+1}^u-\theta_t^u)^2
&\leq\frac{\nu^2T}{4m^3}\\
&\leq\frac{\nu^2T}{4m_0^3}
=W.
\end{align*}
Since the construction also ensures
$\theta_1^u=0$, all the sequences
$\theta^u$ belong to $\Theta_T(W)$.

Moreover, $m\leq T/2$ implies
\begin{align*}
K=\left\lfloor\frac Tm\right\rfloor
\geq\frac{T}{2m}.
\end{align*}

We can now complete the lower bound.
Since all the constructed sequences belong
to $\Theta_T(W)$, the maximum expected
estimation error over this class is at
least the average error over the $2^K$
constructed sequences. Therefore,
\begin{align*}
&\sup_{\theta\in\Theta_T(W)}
\E_\theta\norm{\widehat\theta-\theta}_2^2\\
&\quad\geq
\frac1{2^K}
\sum_{u\in\{-1,+1\}^K}
\E_u\norm{\widehat\theta-\theta^u}_2^2\\
&\quad\geq
\frac{K\nu^2}{768}\\
&\quad\geq
\frac{\nu^2T}{1536m}\\
&\quad\geq
\frac{\nu^2T}{4608m_0}\\
&\quad=
\frac{4^{1/3}}{4608}
(\nu^2T)^{2/3}W^{1/3}\\
&\quad\geq
\frac1{4096}
(\nu^2T)^{2/3}W^{1/3}.
\end{align*}

This bound holds for every estimator
$\widehat\theta$. Taking the infimum
over estimators and substituting
$\nu^2=\sigma^2/(q\kappa^2)$ gives
\begin{align*}
&\inf_{\widehat\theta}
\sup_{\theta\in\Theta_T(W)}
\E_\theta\norm{\widehat\theta-\theta}_2^2\\
&\quad\geq
\frac1{4096}
\frac{(\sigma^2T/q)^{2/3}W^{1/3}}
{\kappa^{4/3}}.
\end{align*}
By the reduction in
Section~\ref{app:gaussian-reduction}, the same
lower bound applies to the original oracle model.
This proves Theorem~\ref{thm:lower} with
$c=1/4096$.
\hfill$\square$

\begin{proof}[Proof of Corollary~\ref{cor:overlap}]
At each update, the learner averages $q$ independent
oracle responses. Since each response has noise
variance $\sigma^2$, the averaged response has
noise variance $\sigma^2/q$. We can therefore
apply Theorem~\ref{thm:upper} with $\tau=1$
and noise parameter $\sigma^2/q$.

Recall that $x_1=\theta_1=0$ and that every
$\theta\in\Theta_T(W)$ satisfies $W_T\leq W$.
Thus, Theorem~\ref{thm:upper} gives, for any
$\eta\in(0,1]$,
\begin{align*}
&\sup_{\theta\in\Theta_T(W)}
\E_\theta\sum_{t=1}^T(x_t-\theta_t)^2\\
&\quad\leq
\frac{\eta\sigma^2T}
{q\kappa(2-\eta\kappa)}
+\frac{W}{\eta^2\kappa^2}\\
&\quad\leq
\frac{\eta\sigma^2T}{q\kappa}
+\frac{W}{\eta^2\kappa^2},
\end{align*}
where the last inequality follows from
$2-\eta\kappa\geq1$.

The first term increases with $\eta$, while
the second decreases with $\eta$. To balance
these terms, we choose the stepsize that
minimizes the last expression. Differentiating
with respect to $\eta$ and setting the
derivative to zero gives
\begin{align*}
\frac{\sigma^2T}{q\kappa}
-\frac{2W}{\eta^3\kappa^2}=0,
\end{align*}
whose solution is
\begin{align*}
\eta_{q,\rm bal}
=\left(\frac{2qW}{\kappa\sigma^2T}\right)^{1/3}.
\end{align*}
Under the assumed condition
$\eta_{q,\rm bal}\leq1$, this is a valid
stepsize. Substituting it into the upper
bound yields
\begin{align*}
&\sup_{\theta\in\Theta_T(W)}
\E_\theta\sum_{t=1}^T(x_t-\theta_t)^2\\
&\quad\leq
\left(2^{1/3}+2^{-2/3}\right)
\frac{(\sigma^2T/q)^{2/3}W^{1/3}}
{\kappa^{4/3}}\\
&\quad<
2\frac{(\sigma^2T/q)^{2/3}W^{1/3}}
{\kappa^{4/3}},
\end{align*}
which proves the claimed tracking guarantee.

It remains to identify the range of $W$
over which this upper bound matches the
lower bound in Theorem~\ref{thm:lower}.
Recall that the lower bound holds for
$T\geq16$ whenever
\begin{align*}
\frac{16\nu^2}{T^2}
\leq W\leq\frac{\nu^2T}{32},
\end{align*}
where $\nu^2=\sigma^2/(q\kappa^2)$.

On the other hand, our chosen stepsize
satisfies $\eta_{q,\rm bal}\leq1$ if and
only if
\begin{align*}
W
&\leq\frac{\kappa\sigma^2T}{2q}\\
&=\frac{\kappa^3\nu^2T}{2}.
\end{align*}
Combining this condition with the variation
range in Theorem~\ref{thm:lower}, we obtain
\begin{align*}
\frac{16\nu^2}{T^2}
\leq W
\leq
\min\left\{
\frac{\nu^2T}{32},
\frac{\kappa^3\nu^2T}{2}
\right\}.
\end{align*}

For $T\geq16$, the first upper limit is
always at least the lower endpoint. The
second is at least the lower endpoint
whenever $\kappa T\geq32^{1/3}$.
Within this range, the upper bound
established above and the lower bound
in Theorem~\ref{thm:lower} agree up to
universal constant factors.
\end{proof}

\subsection{Lower bound with correlated Gaussian noise}
\label{app:correlated-lower}

\begin{proof}[Proof of Theorem~\ref{thm:correlated-lower}]
As in the independent-noise case, we begin by removing
the known contribution of the query locations. The
transformed observations satisfy
\begin{align*}
Z_{t,i}
=\theta_t+\frac{\varepsilon_{t,i}}{\kappa},
\qquad t=1,\ldots,T,\quad i=1,\ldots,q.
\end{align*}
Unlike the previous setting, the observation errors
are now correlated across time.

Recall that
$\chi=(\tau-1)/(\tau+1)$, and define the
$T\times T$ matrix $R_\chi$ by
\begin{align*}
(R_\chi)_{st}=\chi^{|s-t|}.
\end{align*}
For each query index $i$, the observation vector
$Z^{(i)}=(Z_{1,i},\ldots,Z_{T,i})$ is Gaussian
with mean $\theta=(\theta_1,\ldots,\theta_T)$
and covariance matrix
$(\sigma^2/\kappa^2)R_\chi$.
The $q$ observation vectors are independent of
one another.

In this proof, let $\Pp_\theta$ denote the joint
law of $(Z^{(1)},\ldots,Z^{(q)})$ under target
sequence $\theta$. For the family $\theta^u$
constructed below, write
$\Pp_u=\Pp_{\theta^u}$ and let $\E_u$ denote
expectation under this law.

Consider two possible target sequences
$\theta,\theta'\in\R^T$, and let
$v=\theta-\theta'$.
Since the two Gaussian models have the same
covariance matrix, the KL divergence between
their joint observation distributions is
\begin{align*}
D_{\rm KL}(\Pp_\theta\Vert\Pp_{\theta'})
&=\frac{q\kappa^2}{2\sigma^2}
v^\top R_\chi^{-1}v\\
&=\frac{1}{2\nu^2}
v^\top R_\chi^{-1}v,
\end{align*}
where $\nu^2=\sigma^2/(q\kappa^2)$.
We will use this expression to measure how
difficult it is to distinguish two target
sequences.

To bound the KL divergence, we need an
expression for the quadratic form
$v^\top R_\chi^{-1}v$.
A centered Gaussian vector with covariance
matrix $R_\chi$ can be generated by
\begin{align*}
G_1&\sim\mathcal N(0,1),\\
G_t&=\chi G_{t-1}+\sqrt{1-\chi^2}\,U_t,
\qquad t=2,\ldots,T,
\end{align*}
where $G_1,U_2,\ldots,U_T$ are independent
standard normal random variables.
Indeed, this construction gives
$\E[G_sG_t]=\chi^{|s-t|}$.

Consider the joint probability density of the Gaussian
vector $G=(G_1,\ldots,G_T)^\top$. Since
$G\sim\mathcal N(0,R_\chi)$, its density is
\begin{align*}
f_G(g)
=
\frac{1}{(2\pi)^{T/2}\det(R_\chi)^{1/2}}
\exp\left(-\frac12 g^\top R_\chi^{-1}g\right),
\end{align*}
where $g=(g_1,\ldots,g_T)^\top\in\R^T$.

We can also compute this density directly from
the recursive construction of $G$. In particular,
$G_1\sim\mathcal N(0,1)$, and for $t\geq2$,
\begin{align*}
G_t\mid G_{t-1}
\sim\mathcal N(\chi G_{t-1},1-\chi^2).
\end{align*}
Because the recursion defines a Markov process,
the joint density factors as
\begin{align*}
f_G(g)
=f_{G_1}(g_1)
\prod_{t=2}^T
f_{G_t\mid G_{t-1}}(g_t\mid g_{t-1}).
\end{align*}
Substituting the corresponding Gaussian densities
gives
\begin{align*}
f_G(g)
&=\frac{1}{(2\pi)^{T/2}(1-\chi^2)^{(T-1)/2}}\\
&\quad\times
\exp\left(
-\frac12 g_1^2
-\frac{1}{2(1-\chi^2)}
\sum_{t=2}^T(g_t-\chi g_{t-1})^2
\right).
\end{align*}

These are two expressions for the same Gaussian
density. Comparing the quadratic terms in their
exponents, we obtain, for every $v\in\R^T$,
\begin{align*}
v^\top R_\chi^{-1}v
=
v_1^2+
\frac{1}{1-\chi^2}
\sum_{t=2}^T(v_t-\chi v_{t-1})^2.
\end{align*}
This identity expresses the inverse covariance
quadratic form in terms of consecutive differences,
which will be useful for bounding the KL divergence
between the constructed target sequences.

For $v_1=v_T=0$, expanding the squares gives
\begin{align*}
v^\top R_\chi^{-1}v
&=\frac{1}{1-\chi^2}
\sum_{t=2}^T(v_t-\chi v_{t-1})^2\\
&=\frac{1}{1-\chi^2}
\sum_{t=2}^T
\left(v_t^2+\chi^2v_{t-1}^2
-2\chi v_tv_{t-1}\right)\\
&=\frac{1+\chi^2}{1-\chi^2}\norm{v}_2^2
-\frac{2\chi}{1-\chi^2}
\sum_{t=1}^{T-1}v_{t+1}v_t,
\end{align*}
where the last equality uses $v_1=v_T=0$,
so that both sums of squared coordinates equal
$\norm{v}_2^2$.

To simplify the cross-product term, observe that
\begin{align*}
&\sum_{t=1}^{T-1}(v_{t+1}-v_t)^2\\
&\quad=
\sum_{t=1}^{T-1}
\left(v_{t+1}^2+v_t^2-2v_{t+1}v_t\right)\\
&\quad=
2\norm{v}_2^2
-2\sum_{t=1}^{T-1}v_{t+1}v_t.
\end{align*}
Rearranging, we obtain
\begin{align*}
\sum_{t=1}^{T-1}v_{t+1}v_t
=
\norm{v}_2^2
-\frac12\sum_{t=1}^{T-1}(v_{t+1}-v_t)^2.
\end{align*}

Substituting this identity into the previous
expression yields
\begin{align*}
v^\top R_\chi^{-1}v
&=\frac{1+\chi^2}{1-\chi^2}\norm{v}_2^2\\
&\quad-\frac{2\chi}{1-\chi^2}
\left(
\norm{v}_2^2
-\frac12\sum_{t=1}^{T-1}(v_{t+1}-v_t)^2
\right)\\
&=\frac{(1-\chi)^2}{1-\chi^2}\norm{v}_2^2
+\frac{\chi}{1-\chi^2}
\sum_{t=1}^{T-1}(v_{t+1}-v_t)^2\\
&=\frac{1-\chi}{1+\chi}\norm{v}_2^2
+\frac{\chi}{1-\chi^2}
\sum_{t=1}^{T-1}(v_{t+1}-v_t)^2\\
&=\frac1\tau\norm{v}_2^2
+\frac{\tau^2-1}{4\tau}
\sum_{t=1}^{T-1}(v_{t+1}-v_t)^2,
\end{align*}
where the last equality follows from
$\chi=(\tau-1)/(\tau+1)$. This identity is useful because it expresses the KL
divergence in terms of the squared norm of $v$ and
the sum of its squared increments. Both quantities
are controlled by Lemma~\ref{lem:triangle-geometry}
when $v$ is the difference between two target
sequences that differ in the sign of one triangular
block. In particular, the squared norm is bounded
by a quantity proportional to $m$, while the sum
of squared increments is bounded by a quantity
proportional to $1/m$. As we show below, choosing
the block length $m$ sufficiently large relative
to the correlation parameter $\tau$ allows us
to keep the KL divergence small. We can therefore
use the same triangular construction as in the
independent-noise case, with a different choice
of block length and amplitude.

We now construct a family of target sequences
using the triangular sequence from
Section~\ref{app:triangular-packing}. The block length
must be chosen to account for the temporal
correlation. Define
\begin{align*}
m_0
=\left(\frac{\tau\nu^2T}{16W}\right)^{1/3},
\end{align*}
and let $m$ be the smallest odd integer
satisfying $m\geq\max\{5,m_0\}$.

The assumed range
\begin{align*}
\frac{4\tau\nu^2}{T^2}
\leq W\leq\frac{\nu^2T}{128\tau^2}
\end{align*}
implies
\begin{align*}
2\tau\leq m_0\leq\frac T4.
\end{align*}
As in the proof of Theorem~\ref{thm:lower},
the choice of $m$ then gives
\begin{align*}
m_0\leq m\leq3m_0,
\qquad m\leq\frac T2.
\end{align*}
In particular, $m\geq\tau$. Moreover,
the number of blocks satisfies
\begin{align*}
K=\left\lfloor\frac Tm\right\rfloor
\geq\frac{T}{2m}.
\end{align*}

Using this block length, construct the
triangular vectors $\psi^{(j)}$ as before.
For each sign vector
$u\in\{-1,+1\}^K$, define
\begin{align*}
\theta^u
=a\sum_{j=1}^K u_j\psi^{(j)},
\qquad
a^2=\frac{\tau\nu^2}{128m}.
\end{align*}
The construction is identical to the
independent-noise case, except for the
choice of $a$. Each block starts and
ends at zero, so $\theta_1^u=0$.

We first verify that these target sequences
satisfy the variation constraint.
By Lemma~\ref{lem:triangle-geometry},
\begin{align*}
&\sum_{t=1}^{T-1}
(\theta_{t+1}^u-\theta_t^u)^2\\
&\quad=
Ka^2\sum_{i=1}^{m-1}
(\psi_{i+1}-\psi_i)^2\\
&\quad\leq
Ka^2\frac8m\\
&\quad\leq
\frac Tm\frac{\tau\nu^2}{128m}
\frac8m\\
&\quad=
\frac{\tau\nu^2T}{16m^3}\\
&\quad\leq W,
\end{align*}
where the last inequality uses $m\geq m_0$.
Thus, every constructed sequence belongs
to $\Theta_T(W)$.

Next, consider two sign vectors $u$ and
$u'$ that differ in exactly one coordinate
$j$. Their corresponding target sequences
agree outside block $\mathcal I_j$ and
have opposite signs on that block.
Therefore, their difference is
\begin{align*}
v=\theta^u-\theta^{u'}
=\pm2a\psi^{(j)}.
\end{align*}
Since the triangular signal begins and
ends at zero, we have $v_1=v_T=0$.
We can therefore apply the quadratic-form
identity derived above.

By Lemma~\ref{lem:triangle-geometry},
\begin{align*}
\norm{v}_2^2
&=4a^2\norm{\psi}_2^2
\leq4a^2m,
\end{align*}
and
\begin{align*}
\sum_{t=1}^{T-1}(v_{t+1}-v_t)^2
&=4a^2\sum_{i=1}^{m-1}
(\psi_{i+1}-\psi_i)^2\\
&\leq\frac{32a^2}{m}.
\end{align*}
Substituting these bounds into the
quadratic-form identity gives
\begin{align*}
v^\top R_\chi^{-1}v
&\leq
4a^2\left(
\frac m\tau+\frac{2\tau}{m}
\right).
\end{align*}
Consequently,
\begin{align*}
D_{\rm KL}(\Pp_u\Vert\Pp_{u'})
&=\frac{v^\top R_\chi^{-1}v}{2\nu^2}\\
&\leq
\frac{2a^2}{\nu^2}
\left(\frac m\tau+\frac{2\tau}{m}\right)\\
&\leq
\frac{6a^2m}{\tau\nu^2}\\
&=\frac3{64}
<\frac1{16},
\end{align*}
where the second inequality uses $m\geq\tau$,
and the equality follows from
$a^2=\tau\nu^2/(128m)$.

We can now apply the testing argument in
Section~\ref{app:testing-reduction}. Recall that any
estimator $\widehat\theta$ of the target
sequence can be used to estimate the sign
of each triangular block. If the sign on
block $j$ is estimated incorrectly, then
the squared estimation error on that block
is at least $a^2\norm{\psi}_2^2$.

For any two sign vectors that differ only
on block $j$, Pinsker's inequality and the
KL bound above give
\begin{align*}
\operatorname{TV}(\Pp_u,\Pp_{u'})
&\leq
\sqrt{\frac12
D_{\rm KL}(\Pp_u\Vert\Pp_{u'})}\\
&<\frac14.
\end{align*}
Thus, the average probability of incorrectly
identifying the sign on block $j$ under
these two possibilities is at least $1/4$.
Since this holds for every choice of the
other signs, the same lower bound holds
when we average uniformly over all $2^K$
sign vectors.

As in the preceding testing argument,
we obtain
\begin{align*}
&\frac1{2^K}
\sum_{u\in\{-1,+1\}^K}
\E_u\norm{\widehat\theta-\theta^u}_2^2\\
&\quad\geq
\frac K4 a^2\norm{\psi}_2^2\\
&\quad\geq
\frac K4\frac{\tau\nu^2}{128m}\frac m6\\
&\quad=
\frac{K\tau\nu^2}{3072},
\end{align*}
where the second inequality follows from
$\norm{\psi}_2^2\geq m/6$ in
Lemma~\ref{lem:triangle-geometry}.

All the constructed target sequences belong
to $\Theta_T(W)$. Therefore, the worst-case
estimation error over this class is at
least their average estimation error.
Using $K\geq T/(2m)$ and $m\leq3m_0$,
we conclude that
\begin{align*}
\sup_{\theta\in\Theta_T(W)}
\E_\theta\norm{\widehat\theta-\theta}_2^2&\geq
\frac{K\tau\nu^2}{3072}\\
&\geq
\frac{\tau\nu^2T}{6144m}\\
&\geq
\frac{\tau\nu^2T}{18432m_0}\\
&=
\frac{16^{1/3}}{18432}
(\tau\nu^2T)^{2/3}W^{1/3}\\
&\geq
\frac1{8192}
(\tau\nu^2T)^{2/3}W^{1/3}.
\end{align*}
This bound holds for every estimator,
including those that use all observations
from all $T$ rounds. Taking the infimum
over estimators and substituting
$\nu^2=\sigma^2/(q\kappa^2)$ yields
\begin{align*}
&\inf_{\widehat\theta}
\sup_{\theta\in\Theta_T(W)}
\E_\theta\norm{\widehat\theta-\theta}_2^2\\
&\quad\geq
\frac1{8192}
\frac{(\tau\sigma^2T/q)^{2/3}W^{1/3}}
{\kappa^{4/3}},
\end{align*}
which proves the claimed lower bound.

For completeness, we verify the range in
which this lower bound matches the tracking
upper bound. Let
\begin{align*}
\bar\varepsilon_t
=\frac1q\sum_{i=1}^q\varepsilon_{t,i}
\end{align*}
denote the averaged oracle noise.
Since the noise sequences for different
query indices are independent,
\begin{align*}
\E[\bar\varepsilon_t]&=0,\\
\E[\bar\varepsilon_s\bar\varepsilon_t]
&=\frac{\sigma^2}{q}\chi^{|s-t|}.
\end{align*}
By Proposition~\ref{prop:geometric-correlation},
the averaged noise satisfies
Assumption~\ref{ass:covariance} with
parameters $(\sigma^2/q,\tau)$.

We may therefore apply
Theorem~\ref{thm:upper} with $x_1=0$
and $W_T\leq W$. Choosing
\begin{align*}
\eta
=\left(
\frac{2qW}{\kappa\tau\sigma^2T}
\right)^{1/3}
\end{align*}
and balancing the noise and variation
terms as in Corollary~\ref{cor:tuned}
gives
\begin{align*}
&\sup_{\theta\in\Theta_T(W)}
\E_\theta\sum_{t=1}^T(x_t-\theta_t)^2\\
&\quad<
2(\tau\nu^2T)^{2/3}W^{1/3},
\end{align*}
provided $\eta\leq1$.

The stepsize condition is equivalent to
\begin{align*}
W\leq\frac{\kappa^3\tau\nu^2T}{2}.
\end{align*}
Combining this with the variation range
in Theorem~\ref{thm:correlated-lower},
the upper and lower bounds agree up to
universal constants whenever
\begin{align*}
\frac{4\tau\nu^2}{T^2}
\leq W
\leq
\min\left\{
\frac{\nu^2T}{128\tau^2},
\frac{\kappa^3\tau\nu^2T}{2}
\right\}.
\end{align*}
Under the theorem's assumption
$\tau\leq T/8$, this interval is
nonempty whenever $\kappa T\geq2$.
\end{proof}

\section{Proof of the adaptive result}
\label{app:meta}

We prove the adaptive guarantee by comparing the
squared distances between the trackers' predictions
and the validation observations $Y_t$. Since these
observations are conditionally unbiased estimates
of the true fixed points, this comparison also
provides a bound on the expected tracking error.
We then apply Theorem~\ref{thm:upper} to compare
the trackers in our stepsize grid with any
constant stepsize in $[1/T,1]$.

\begin{proof}[Proof of Theorem~\ref{thm:meta}]
For each expert $j=0,\ldots,J$, define its
validation loss at round $t$ as
\begin{align*}
\ell_{t,j}
=\norm{\widetilde x_{t,j}-Y_t}^2.
\end{align*}
Similarly, the loss of the combined prediction
$\bar x_t$ is
\begin{align*}
\bar\ell_t=\norm{\bar x_t-Y_t}^2.
\end{align*}
Recall that
$\bar x_t=\sum_{j=0}^J p_{t,j}\widetilde x_{t,j}$,
where the weights are nonnegative and sum to one.
Since all predictions and $Y_t$ belong to $\cX$,
Lemma~\ref{lem:exp-concavity} gives
\begin{align*}
\bar\ell_t
\leq
-\frac1\lambda
\log\left(
\sum_{j=0}^J p_{t,j}
e^{-\lambda\ell_{t,j}}
\right),
\end{align*}
where $\lambda=1/(8B^2)$.

To relate this inequality to the cumulative
losses, define
\begin{align*}
Z_t
=
\sum_{j=0}^J p_{1,j}
\exp\left(
-\lambda\sum_{s=1}^{t-1}\ell_{s,j}
\right),
\qquad t=1,\ldots,T+1.
\end{align*}
Since the initial weights are uniform,
$Z_1=1$. The exponential-weighting update
in Algorithm~\ref{alg:meta-track} implies
\begin{align*}
p_{t,j}
=
\frac{
p_{1,j}\exp\left(
-\lambda\sum_{s=1}^{t-1}\ell_{s,j}
\right)
}{Z_t}.
\end{align*}
Consequently,
\begin{align*}
\sum_{j=0}^J p_{t,j}e^{-\lambda\ell_{t,j}}
&=
\frac1{Z_t}
\sum_{j=0}^J p_{1,j}
\exp\left(
-\lambda\sum_{s=1}^t\ell_{s,j}
\right)\\
&=\frac{Z_{t+1}}{Z_t}.
\end{align*}
Substituting this identity into the bound
on $\bar\ell_t$ yields
\begin{align*}
\bar\ell_t
\leq-\frac1\lambda
\log\left(\frac{Z_{t+1}}{Z_t}\right).
\end{align*}
Summing over $t=1,\ldots,T$, the logarithmic
terms telescope, giving
\begin{align*}
\sum_{t=1}^T\bar\ell_t
&\leq
-\frac1\lambda
\log\left(\frac{Z_{T+1}}{Z_1}\right)\\
&=-\frac1\lambda\log Z_{T+1}.
\end{align*}
Here, $Z_{T+1}$ is defined only for the
analysis; the algorithm does not need to
update its weights after the final round.

For any fixed expert $j$, the definition
of $Z_{T+1}$ and the uniform initial
weights imply
\begin{align*}
Z_{T+1}
&=\frac1{J+1}
\sum_{k=0}^J
\exp\left(
-\lambda\sum_{t=1}^T\ell_{t,k}
\right)\\
&\geq
\frac1{J+1}
\exp\left(
-\lambda\sum_{t=1}^T\ell_{t,j}
\right).
\end{align*}
Taking logarithms and combining this
with the previous inequality gives
\begin{align*}
\sum_{t=1}^T\bar\ell_t
\leq
\sum_{t=1}^T\ell_{t,j}
+\frac{\log(J+1)}{\lambda}.
\end{align*}
This inequality holds for every expert $j$
and every realization of the validation
observations. Thus, the cumulative validation
loss of the combined predictor is no larger
than that of the best individual expert,
apart from the additional term
$\lambda^{-1}\log(J+1)$.

We now relate the validation losses to the
actual tracking errors. Write
\begin{align*}
Y_t=x_t^*+\varepsilon_t^{\rm val},
\end{align*}
where
$\varepsilon_t^{\rm val}=Y_t-x_t^*$.
By Assumption~\ref{ass:validation},
\begin{align*}
\E[\varepsilon_t^{\rm val}\mid\cG_t]=0.
\end{align*}
For any $\cG_t$-measurable prediction $x$,
expanding the squared norm gives
\begin{align*}
\norm{x-Y_t}^2
&=\norm{x-x_t^*}^2
+\norm{\varepsilon_t^{\rm val}}^2\\
&\quad
-2\ip{x-x_t^*}{\varepsilon_t^{\rm val}}.
\end{align*}
Taking conditional expectations, the cross
term vanishes because $x$ is
$\cG_t$-measurable and the validation
error has conditional mean zero. Hence,
\begin{align*}
\E[\norm{x-Y_t}^2\mid\cG_t]
&=\norm{x-x_t^*}^2\\
&\quad+
\E[\norm{\varepsilon_t^{\rm val}}^2
\mid\cG_t].
\end{align*}

This identity applies both to the combined
prediction $\bar x_t$ and to each expert's
prediction $\widetilde x_{t,j}$. In both cases,
the expected validation loss equals the
expected squared tracking error plus the
same validation-noise term.

We can therefore take expectations in the
cumulative validation-loss inequality and
cancel the common validation-noise terms.
For every expert $j$, this gives
\begin{align*}
&\sum_{t=1}^T
\E\norm{\bar x_t-x_t^*}^2\\
&\quad\leq
\sum_{t=1}^T
\E\norm{\widetilde x_{t,j}-x_t^*}^2
+\frac{\log(J+1)}{\lambda}.
\end{align*}
Recall that
$\widetilde x_{t,j}=\Pi_{\cX}(x_{t,j})$.
Since $\cX$ is closed and convex and
$x_t^*\in\cX$, projection cannot increase
the distance to $x_t^*$. Thus,
\begin{align*}
\norm{\widetilde x_{t,j}-x_t^*}
\leq\norm{x_{t,j}-x_t^*}.
\end{align*}
Taking the minimum over $j=0,\ldots,J$
and substituting $\lambda=1/(8B^2)$,
we obtain
\begin{align*}
&\sum_{t=1}^T
\E\norm{\bar x_t-x_t^*}^2\\
&\quad\leq
\min_{0\leq j\leq J}
\sum_{t=1}^T
\E\norm{x_{t,j}-x_t^*}^2
+8B^2\log(J+1),
\end{align*}
which proves~\eqref{eq:meta-grid}.

We next establish~\eqref{eq:meta-bound} by
comparing the stepsizes in our grid with
an arbitrary constant stepsize.
Fix any $\eta\in[1/T,1]$. Since the grid
contains the stepsizes
$\eta_j=2^{-j}$ for $j=0,\ldots,J$, with
$J=\lceil\log_2T\rceil$, there is an
index $j$ such that
\begin{align*}
\frac{\eta}{2}\leq\eta_j\leq\eta.
\end{align*}

Applying Theorem~\ref{thm:upper} to
expert $j$ gives
\begin{align*}
&\sum_{t=1}^T
\E\norm{x_{t,j}-x_t^*}^2\\
&\quad\leq
\frac{\norm{x_1-x_1^*}^2}{\eta_j\kappa}
+\frac{\eta_j\tau\sigma^2T}
{\kappa(2-\eta_j\kappa)}
+\frac{W_T}{\eta_j^2\kappa^2}.
\end{align*}
We now express the right-hand side in
terms of $\eta$. Since $\eta_j\geq\eta/2$,
the initialization term increases by at
most a factor of two, and the variation
term by at most a factor of four.
Moreover, the function
$u\mapsto u/(2-\kappa u)$ is increasing
on $(0,1]$, so $\eta_j\leq\eta$ implies
\begin{align*}
\frac{\eta_j}{2-\eta_j\kappa}
\leq\frac{\eta}{2-\eta\kappa}.
\end{align*}
Combining these observations, we obtain
\begin{align*}
&\sum_{t=1}^T
\E\norm{x_{t,j}-x_t^*}^2\\
&\quad\leq
\frac{2\norm{x_1-x_1^*}^2}{\eta\kappa}
+\frac{\eta\tau\sigma^2T}
{\kappa(2-\eta\kappa)}
+\frac{4W_T}{\eta^2\kappa^2}\\
&\quad\leq
4\left\{
\frac{\norm{x_1-x_1^*}^2}{\eta\kappa}
+\frac{\eta\tau\sigma^2T}
{\kappa(2-\eta\kappa)}
+\frac{W_T}{\eta^2\kappa^2}
\right\}.
\end{align*}

Since~\eqref{eq:meta-grid} compares the
combined predictor with the best expert
in the grid, the same upper bound applies
to the combined predictor, together with
the additional term $8B^2\log(J+1)$.
The inequality holds for every
$\eta\in[1/T,1]$, so taking the infimum
over this interval proves
\eqref{eq:meta-bound}.

Finally, suppose $x_1=x_1^*$,
$W_T>0$, $\tau\sigma^2>0$, and the
balancing stepsize $\eta_{\rm bal}$
from Corollary~\ref{cor:tuned} lies
in $[1/T,1]$. Applying
\eqref{eq:meta-bound} with this stepsize
and using Corollary~\ref{cor:tuned}
gives
\begin{align*}
&\sum_{t=1}^T
\E\norm{\bar x_t-x_t^*}^2\\
&\quad<
8\frac{(\tau\sigma^2T)^{2/3}W_T^{1/3}}
{\kappa^{4/3}}
+8B^2\log(1+\lceil\log_2T\rceil),
\end{align*}
which completes the proof.
\end{proof}

\paragraph{Proof of the bound for direct smoothing.}
We now consider the simpler recursion
\begin{align*}
u_{t+1}=(1-a)u_t+aY_t,
\qquad u_1=x_1,
\end{align*}
where $a\in(0,1]$. Unlike the adaptive
algorithm, this recursion uses only the
validation observations and does not
require operator evaluations.

We first bound the variance of the
validation observations. Let $c$ be the
center of a ball of radius $B$ containing
$\cX$. Since $Y_t\in\cX$ and
$\E[Y_t\mid\cG_t]=x_t^*$, the conditional
variance identity gives
\begin{align*}
&\E[\norm{Y_t-x_t^*}^2\mid\cG_t]\\
&\quad=
\E[\norm{Y_t-c}^2\mid\cG_t]
-\norm{x_t^*-c}^2\\
&\quad\leq B^2.
\end{align*}
Thus, the validation errors
$Y_t-x_t^*$ have conditional mean zero
and conditional second moments bounded
by $B^2$.

To apply Theorem~\ref{thm:upper},
consider the time-varying constant map
\begin{align*}
F_t(u)=x_t^*.
\end{align*}
This map has fixed point $x_t^*$ and
contraction gap equal to one. The
direct-smoothing recursion can be
written as
\begin{align*}
u_{t+1}
=(1-a)u_t
+a\left(F_t(u_t)+Y_t-x_t^*\right).
\end{align*}
Because the validation errors have
conditional mean zero given the full
past, they form a martingale-difference
sequence. Consequently, the noise
assumption in Theorem~\ref{thm:upper}
holds with $\sigma^2=B^2$ and $\tau=1$.

Applying Theorem~\ref{thm:upper}
with stepsize $a$, contraction gap one,
and initial iterate $u_1=x_1$, we obtain
\begin{align*}
\sum_{t=1}^T\E\norm{u_t-x_t^*}^2
&\leq
\frac{\norm{x_1-x_1^*}^2}{a}
+\frac{aB^2T}{2-a}
+\frac{W_T}{a^2}.
\end{align*}
This proves the direct-smoothing bound
stated in Section~\ref{sec:meta}.

To compare this bound with the operator
tracking guarantee in
Theorem~\ref{thm:upper}, set
$a=\eta\kappa$. Under this choice,
both bounds have the same initialization
and target-movement terms. Their noise
terms are, respectively,
\begin{align*}
\frac{aB^2T}{2-a}
\qquad\text{and}\qquad
\frac{a\tau\sigma^2T}{\kappa^2(2-a)}.
\end{align*}
Thus, the operator-tracking upper bound
has a smaller noise term whenever
$\tau\sigma^2/\kappa^2<B^2$, as discussed
in Section~\ref{sec:meta}.
\hfill$\square$

\section{Application to tabular policy evaluation}
\label{app:td}

\subsection{Model and guarantees}

We now apply our tracking results to tabular policy
evaluation when the rewards and transition probabilities
change over time. We consider a deterministic sequence
of discounted Markov reward processes
$(\cS,P_t,r_t,\gamma)$, where $\cS$ is a finite
state space, $P_t$ is the transition matrix,
$r_t$ is the expected reward vector, and
$\gamma\in[0,1)$ is the discount factor.
The sequence of Markov reward processes is fixed
before learning begins.

We assume access to a generative model that can
sample rewards and next states from any state.
At each round $t$, the current Markov reward
process remains fixed while the learner makes
its queries. A query at state $s\in\cS$ returns
a reward $R$ and a next state $S'$ satisfying
\begin{align*}
\abs{R}&\leq R_{\max}\quad\text{a.s.},\\
\E[R]&=r_t(s),\\
\operatorname{Var}(R)&\leq\sigma_R^2,\\
S'&\sim P_t(\cdot\mid s).
\end{align*}
Each simulator call returns a new sample $(R,S')$
based on the queried state $s$ and the current
Markov reward process. Conditional on the queried
state and all previous observations, the sample
follows the reward and transition distributions
specified above. Different calls use independent
randomness, although the reward and next state
returned by the same call may be dependent. In particular, the reward may depend on the next state. For
example, transitions to different states may
produce different rewards.
During a rollout, the next queried state is
determined by the previous transition, so the
sequence of visited states is generally dependent.

We further assume that all transition matrices
share a common invariant distribution $\mu$,
with $\mu(s)>0$ for every $s\in\cS$. Thus,
\begin{align*}
\mu P_t=\mu,
\qquad t=1,\ldots,T.
\end{align*}
This assumption allows us to use a common norm
in which the Bellman operators are contractions.
In particular, define
\begin{align*}
\norm{v}_\mu^2
&=\sum_{s\in\cS}\mu(s)v(s)^2,\\
\norm{A}_{\mu\to\mu}
&=\sup_{v\neq0}\frac{\norm{Av}_\mu}{\norm{v}_\mu}.
\end{align*}
The common invariant distribution restricts
how the transition matrices may change, but
does not require them to remain fixed.

At round $t$, the Bellman operator is
\begin{align*}
\cT_tv=r_t+\gamma P_tv,
\end{align*}
and its unique fixed point is
\begin{align*}
v_t^*=(I-\gamma P_t)^{-1}r_t.
\end{align*}
Our goal is to track this changing value
function.

For synchronous tabular TD, we make one
generative-model call from each state at
every update round. Let $(R_{t,s},S'_{t,s})$
denote the sample obtained from state $s$
at round $t$. These samples define the
random Bellman operator
\begin{align*}
(\widehat\cT_tv)(s)
=R_{t,s}+\gamma v(S'_{t,s}).
\end{align*}
Starting from a deterministic vector $v_1$,
the TD iteration with constant stepsize
$\eta\in(0,1]$ is
\begin{align*}
v_{t+1}
=(1-\eta)v_t+\eta\widehat\cT_tv_t,
\qquad t=1,\ldots,T-1.
\end{align*}

The following corollary applies
Theorem~\ref{thm:upper} to this iteration.

\begin{corollary}[Synchronous tabular TD]
\label{cor:td}
Let $V_{\max}=R_{\max}/(1-\gamma)$ and
suppose $\norm{v_1}_\infty\leq V_{\max}$.
Define
\begin{align*}
\sigma_{\rm TD}^2
&=2\sigma_R^2+2\gamma^2V_{\max}^2,\\
W_T^V
&=\sum_{t=1}^{T-1}
\norm{v_{t+1}^*-v_t^*}_\mu^2.
\end{align*}
Then, for every $\eta\in(0,1]$,
\begin{align*}
&\sum_{t=1}^T\E\norm{v_t-v_t^*}_\mu^2\\
&\quad\leq
\frac{\norm{v_1-v_1^*}_\mu^2}
{\eta(1-\gamma)}
+\frac{\eta\sigma_{\rm TD}^2T}
{(1-\gamma)(2-\eta(1-\gamma))}\\
&\qquad+
\frac{W_T^V}{\eta^2(1-\gamma)^2}.
\end{align*}

Moreover, define the squared variations
of the rewards and transition matrices by
\begin{align*}
W_T^r
&=\sum_{t=1}^{T-1}
\norm{r_{t+1}-r_t}_\mu^2,\\
W_T^P
&=\sum_{t=1}^{T-1}
\norm{P_{t+1}-P_t}_{\mu\to\mu}^2.
\end{align*}
Then
\begin{align*}
W_T^V
\leq
\frac{2}{(1-\gamma)^2}
\left(
W_T^r+\gamma^2V_{\max}^2W_T^P
\right).
\end{align*}
\end{corollary}

The first bound controls the cumulative
value-estimation error in terms of
initialization, sampling noise, and changes
in the true value function. The second
relates the movement of the value function
to changes in rewards and transition
probabilities.

We next consider stepsize adaptation.
The adaptive algorithm in
Section~\ref{sec:meta} requires additional
observations that are unbiased estimates
of the current fixed point. In the present
setting, these are observations of the
value vector $v_t^*$.

We construct such observations using
generative-model rollouts. The idea is to
sample a single reward at a random time,
where the probability of selecting that
time follows the discount factor $\gamma$.
Rescaling the sampled reward then gives
an unbiased estimate of the discounted
value function.

\begin{proposition}[An unbiased validation vector]
\label{prop:validation-td}
Fix a round $t$ and keep its Markov reward
process unchanged throughout the following
construction. For every starting state
$s\in\cS$, independently draw an integer
$L_{t,s}\geq0$ with distribution
\begin{align*}
\Pp(L_{t,s}=\ell)
=(1-\gamma)\gamma^\ell,
\qquad \ell=0,1,\ldots.
\end{align*}
Starting from $s$, follow the transition
matrix $P_t$ for $L_{t,s}$ steps. At the
resulting state, make one additional
generative-model call and let
$R_{t,s}^{\rm val}$ denote its reward.
Define
\begin{align*}
Y_t(s)=\frac{R_{t,s}^{\rm val}}{1-\gamma},
\qquad s\in\cS.
\end{align*}

If the rollouts are independent across
starting states and use samples independent
of those used for the TD updates, then
\begin{align*}
\E[Y_t\mid\cG_t]&=v_t^*,\\
\norm{Y_t}_\mu&\leq V_{\max}
\quad\text{almost surely}.
\end{align*}
The expected number of generative-model
calls required to construct $Y_t$ is
$|\cS|/(1-\gamma)$.
\end{proposition}

The proposition provides the additional
observations required by
Assumption~\ref{ass:validation}. Unlike
ordinary synchronous TD, the adaptive
method needs to know $\mu$, since it uses
the $L_2(\mu)$ norm to compare the
predictions of different trackers.

We can now apply Theorem~\ref{thm:meta}
to obtain an adaptive TD guarantee.

\begin{corollary}[Adaptive synchronous TD]
\label{cor:meta-td}
Suppose the conditions of
Corollary~\ref{cor:td} hold, and assume
that $\mu$ is known and $R_{\max}>0$.
Run Algorithm~\ref{alg:meta-track} in
$L_2(\mu)$ with
\begin{align*}
B&=V_{\max},\\
\cX&=[-V_{\max},V_{\max}]^{|\cS|}.
\end{align*}
At each update round, use the same sampled
Bellman operator for all experts, and
construct the validation observation $Y_t$
as in Proposition~\ref{prop:validation-td}.

Then the combined value estimate $\bar v_t$
satisfies
\begin{align*}
&\sum_{t=1}^T
\E\norm{\bar v_t-v_t^*}_\mu^2\\
&\quad\leq
4\inf_{\eta\in[1/T,1]}
\left\{
\frac{\norm{v_1-v_1^*}_\mu^2}
{\eta(1-\gamma)}
+\frac{\eta\sigma_{\rm TD}^2T}
{(1-\gamma)(2-\eta(1-\gamma))}
+\frac{W_T^V}{\eta^2(1-\gamma)^2}
\right\}\\
&\qquad+
8V_{\max}^2
\log(1+\lceil\log_2T\rceil).
\end{align*}

The TD updates require
$(T-1)|\cS|$ generative-model calls.
The validation observations require
an additional expected
$T|\cS|/(1-\gamma)$ calls.
Thus, validation requires a factor
$1/(1-\gamma)$ more calls per prediction
round than a synchronous TD update.
\end{corollary}

The adaptive guarantee compares the
combined prediction with TD trackers
using different constant stepsizes.
However, the validation observations
can also be used directly, without
running TD updates. In particular,
the direct-smoothing recursion from
Section~\ref{sec:meta} applies with
$B=V_{\max}$ and $W_T=W_T^V$.

Direct smoothing uses the same validation
rollouts but does not require the additional
$(T-1)|\cS|$ calls used for the TD updates.
It is therefore a useful baseline under
this observation model. Our result
establishes the adaptive guarantee
relative to the TD trackers, but does
not establish a sampling advantage
over direct smoothing.

\subsection{Proofs}

\begin{proof}[Proof of Corollary~\ref{cor:td}]
We verify the contraction and noise
conditions of Theorem~\ref{thm:upper}.

First, recall that all transition matrices
share the invariant distribution $\mu$.
For any vector $v\in\R^{|\cS|}$,
Jensen's inequality gives
\begin{align*}
\norm{P_tv}_\mu^2
&=\sum_s\mu(s)
\left(
\sum_{s'}P_t(s'\mid s)v(s')
\right)^2\\
&\leq
\sum_s\mu(s)
\sum_{s'}P_t(s'\mid s)v(s')^2\\
&=
\sum_{s'}v(s')^2
\sum_s\mu(s)P_t(s'\mid s)\\
&=
\sum_{s'}\mu(s')v(s')^2\\
&=\norm{v}_\mu^2.
\end{align*}
The penultimate equality follows from
$\mu P_t=\mu$. Therefore,
$\norm{P_t}_{\mu\to\mu}\leq1$, and
for any two value vectors $v,w$,
\begin{align*}
\norm{\cT_tv-\cT_tw}_\mu
&=\gamma\norm{P_t(v-w)}_\mu\\
&\leq\gamma\norm{v-w}_\mu.
\end{align*}
Thus, the mean Bellman operator is a
contraction in $L_2(\mu)$ with
contraction gap $\kappa=1-\gamma$.

We next show that the TD iterates are
bounded. Consider the set
\begin{align*}
\cX=[-V_{\max},V_{\max}]^{|\cS|}.
\end{align*}
For any $v\in\cX$ and any sampled
transition from state $s$, we have
\begin{align*}
\abs{R_{t,s}+\gamma v(S'_{t,s})}
&\leq R_{\max}+\gamma V_{\max}\\
&=V_{\max}.
\end{align*}
Hence, the sampled Bellman operator maps
$\cX$ into itself. Since the TD update
is a convex combination of $v_t$ and
$\widehat\cT_tv_t$, it also remains in
$\cX$. The assumption $v_1\in\cX$
therefore implies
\begin{align*}
\norm{v_t}_\infty\leq V_{\max},
\qquad t=1,\ldots,T.
\end{align*}
Moreover, because
$\abs{r_t(s)}\leq R_{\max}$,
the true value functions satisfy
\begin{align*}
\norm{v_t^*}_\infty\leq V_{\max}.
\end{align*}

We now bound the observation noise.
Define
\begin{align*}
\xi_{t+1}
=\widehat\cT_tv_t-\cT_tv_t.
\end{align*}
For each state $s$, this noise can be
written as
\begin{align*}
\xi_{t+1}(s)
&=R_{t,s}-r_t(s)\\
&\quad+\gamma\left(
v_t(S'_{t,s})-(P_tv_t)(s)
\right).
\end{align*}
Conditional on $\cF_t$, the current
iterate $v_t$ is fixed, while the
generative-model samples are fresh.
The reward and transition assumptions
therefore give
\begin{align*}
\E[\xi_{t+1}\mid\cF_t]=0.
\end{align*}

To bound the second moment, we use
$(a+b)^2\leq2a^2+2b^2$. This does not
require the reward and next state from
the same call to be independent.
Consequently,
\begin{align*}
&\E[\norm{\xi_{t+1}}_\mu^2\mid\cF_t]\\
&\quad\leq
2\sum_s\mu(s)
\E[(R_{t,s}-r_t(s))^2\mid\cF_t]\\
&\qquad+
2\gamma^2\sum_s\mu(s)
\E\left[
\left(
v_t(S'_{t,s})-(P_tv_t)(s)
\right)^2
\mid\cF_t
\right]\\
&\quad\leq
2\sigma_R^2+
2\gamma^2\sum_s\mu(s)
\E[v_t(S'_{t,s})^2\mid\cF_t]\\
&\quad=
2\sigma_R^2+
2\gamma^2\sum_s\mu(s)
\sum_{s'}P_t(s'\mid s)v_t(s')^2\\
&\quad=
2\sigma_R^2+
2\gamma^2\norm{v_t}_\mu^2\\
&\quad\leq
2\sigma_R^2+2\gamma^2V_{\max}^2\\
&\quad=\sigma_{\rm TD}^2.
\end{align*}
Here, we used the common invariant
distribution in the second-to-last
equality, and the boundedness of
$v_t$ in the last inequality.

Thus, the noise is a martingale-difference
sequence with conditional second moment
bounded by $\sigma_{\rm TD}^2$.
It satisfies Assumption~\ref{ass:covariance}
with $\tau=1$.

We can now apply Theorem~\ref{thm:upper}
with $\kappa=1-\gamma$ and
$\sigma^2=\sigma_{\rm TD}^2$.
This gives
\begin{align*}
&\sum_{t=1}^T
\E\norm{v_t-v_t^*}_\mu^2\\
&\quad\leq
\frac{\norm{v_1-v_1^*}_\mu^2}
{\eta(1-\gamma)}
+\frac{\eta\sigma_{\rm TD}^2T}
{(1-\gamma)(2-\eta(1-\gamma))}\\
&\qquad+
\frac{W_T^V}{\eta^2(1-\gamma)^2},
\end{align*}
which proves the first claim.

It remains to relate the value-function
variation $W_T^V$ to changes in the
rewards and transition matrices.
Recall that the value functions satisfy
the Bellman equations
\begin{align*}
v_t^*&=r_t+\gamma P_tv_t^*,\\
v_{t+1}^*&=r_{t+1}+\gamma P_{t+1}v_{t+1}^*.
\end{align*}
Subtracting these equations and
rearranging gives
\begin{align*}
&(I-\gamma P_{t+1})
(v_{t+1}^*-v_t^*)\\
&\quad=
(r_{t+1}-r_t)
+\gamma(P_{t+1}-P_t)v_t^*.
\end{align*}

Since
$\norm{P_{t+1}}_{\mu\to\mu}\leq1$
and $\gamma<1$, the Neumann series
implies
\begin{align*}
\norm{(I-\gamma P_{t+1})^{-1}}_{\mu\to\mu}
&\leq
\sum_{k=0}^\infty\gamma^k\\
&=\frac{1}{1-\gamma}.
\end{align*}
Multiplying the preceding identity
by $(I-\gamma P_{t+1})^{-1}$
and taking norms, we obtain
\begin{align*}
&\norm{v_{t+1}^*-v_t^*}_\mu\\
&\quad\leq
\frac{1}{1-\gamma}
\left(
\norm{r_{t+1}-r_t}_\mu
+\gamma
\norm{P_{t+1}-P_t}_{\mu\to\mu}
\norm{v_t^*}_\mu
\right)\\
&\quad\leq
\frac{1}{1-\gamma}
\left(
\norm{r_{t+1}-r_t}_\mu
+\gamma V_{\max}
\norm{P_{t+1}-P_t}_{\mu\to\mu}
\right).
\end{align*}
Squaring both sides, using
$(a+b)^2\leq2a^2+2b^2$, and summing
over $t=1,\ldots,T-1$ gives
\begin{align*}
W_T^V
&\leq
\frac{2}{(1-\gamma)^2}
\sum_{t=1}^{T-1}
\left(
\norm{r_{t+1}-r_t}_\mu^2
+\gamma^2V_{\max}^2
\norm{P_{t+1}-P_t}_{\mu\to\mu}^2
\right)\\
&=
\frac{2}{(1-\gamma)^2}
\left(
W_T^r+\gamma^2V_{\max}^2W_T^P
\right),
\end{align*}
as claimed.
\end{proof}

\begin{proof}[Proof of Proposition~\ref{prop:validation-td}]
Fix a round $t$ and a starting state
$s\in\cS$. Let
$S_0^{t,s}=s$ and denote by
$S_\ell^{t,s}$ the state reached
after $\ell$ transitions under $P_t$.

Recall that the random horizon
$L_{t,s}$ has distribution
\begin{align*}
\Pp(L_{t,s}=\ell)
=(1-\gamma)\gamma^\ell,
\qquad \ell=0,1,\ldots.
\end{align*}
The horizon is drawn independently
of the rollout samples, and the
Markov reward process remains fixed
during the rollout.

Conditional on $L_{t,s}=\ell$,
the learner follows $P_t$ for
$\ell$ steps and then samples a
reward at state $S_\ell^{t,s}$.
The conditional mean of this reward
is $r_t(S_\ell^{t,s})$. Therefore,
\begin{align*}
&\E[R_{t,s}^{\rm val}
\mid L_{t,s}=\ell,\cG_t]\\
&\quad=
\E[r_t(S_\ell^{t,s})
\mid S_0^{t,s}=s]\\
&\quad=(P_t^\ell r_t)(s).
\end{align*}

We now compute the conditional expectation
of the validation observation
$Y_t(s)=R_{t,s}^{\rm val}/(1-\gamma)$.
By conditioning on the random horizon,
\begin{align*}
\E[Y_t(s)\mid\cG_t]
&=\frac{1}{1-\gamma}
\sum_{\ell=0}^\infty
\Pp(L_{t,s}=\ell)
\E[R_{t,s}^{\rm val}
\mid L_{t,s}=\ell,\cG_t]\\
&=\frac{1}{1-\gamma}
\sum_{\ell=0}^\infty
(1-\gamma)\gamma^\ell
(P_t^\ell r_t)(s)\\
&=\sum_{\ell=0}^\infty
\gamma^\ell(P_t^\ell r_t)(s)\\
&=[(I-\gamma P_t)^{-1}r_t](s)\\
&=v_t^*(s).
\end{align*}
The fourth equality follows from the
Neumann series, which converges because
$\gamma<1$ and $P_t$ is a transition
matrix.

Since this equality holds for every
starting state $s$, we conclude that
\begin{align*}
\E[Y_t\mid\cG_t]=v_t^*.
\end{align*}
Thus, $Y_t$ is a conditionally unbiased
observation of the current value vector.

Next, the reward bound gives
\begin{align*}
\abs{Y_t(s)}
&=\frac{\abs{R_{t,s}^{\rm val}}}{1-\gamma}\\
&\leq\frac{R_{\max}}{1-\gamma}
=V_{\max}.
\end{align*}
Consequently,
\begin{align*}
\norm{Y_t}_\mu^2
&=\sum_s\mu(s)Y_t(s)^2\\
&\leq V_{\max}^2\sum_s\mu(s)\\
&=V_{\max}^2.
\end{align*}
This proves that
$\norm{Y_t}_\mu\leq V_{\max}$
almost surely.

It remains to compute the sampling cost.
For each starting state, a rollout of
length $L_{t,s}$ uses $L_{t,s}$
transition samples and one additional
call to obtain the final reward.
The expected number of calls is therefore
\begin{align*}
\E[L_{t,s}+1]
&=\sum_{\ell=0}^\infty
(\ell+1)(1-\gamma)\gamma^\ell\\
&=\frac{1}{1-\gamma}.
\end{align*}
Since we construct one validation
observation for each state, the total
expected number of calls per round is
\begin{align*}
\frac{|\cS|}{1-\gamma}.
\end{align*}

Finally, the validation rollouts are
generated independently of the samples
used for the TD updates. Conditional
on $\cG_t$, this verifies the independence
requirement in
Assumption~\ref{ass:validation}.
\end{proof}

\begin{proof}[Proof of Corollary~\ref{cor:meta-td}]
We verify that the conditions of
Theorem~\ref{thm:meta} hold for the
synchronous TD application.

Recall that
$\cX=[-V_{\max},V_{\max}]^{|\cS|}$.
By the proof of
Corollary~\ref{cor:td}, every TD
iterate remains in $\cX$.
The same argument applies to each
expert because all experts start
from $v_1$ and use stepsizes in
$(0,1]$. Thus, projection onto
$\cX$ does not change their iterates.

Moreover, every true value vector
$v_t^*$ belongs to $\cX$.
Since $\mu$ is a probability
distribution, every $v\in\cX$
satisfies
\begin{align*}
\norm{v}_\mu^2
=\sum_s\mu(s)v(s)^2
\leq V_{\max}^2.
\end{align*}
Hence, $\cX$ is contained in the
radius-$V_{\max}$ ball of $L_2(\mu)$,
as required by Theorem~\ref{thm:meta}.

At each update round, one collection
of samples $(R_{t,s},S'_{t,s})$,
$s\in\cS$, defines a random Bellman
operator. The same samples can be
used to evaluate this operator at
the current iterate of every expert.
In particular, the experts do not
need independent collections of
generative-model samples.

For each fixed expert, the noise
remains conditionally mean zero
with second moment bounded by
$\sigma_{\rm TD}^2$, as established
in the proof of
Corollary~\ref{cor:td}.
Thus, each expert satisfies
Assumption~\ref{ass:covariance}
with $\tau=1$ and
$\sigma^2=\sigma_{\rm TD}^2$.
The noises of different experts
may be dependent, which is allowed
by Theorem~\ref{thm:meta}.

Proposition~\ref{prop:validation-td}
also provides a validation vector
$Y_t\in\cX$ satisfying
\begin{align*}
\E[Y_t\mid\cG_t]=v_t^*,
\end{align*}
with validation samples independent
of the TD update samples.
Therefore, Assumption~\ref{ass:validation}
is satisfied with $B=V_{\max}$.

We may now apply
Theorem~\ref{thm:meta} with
\begin{align*}
\kappa&=1-\gamma,\\
\sigma^2&=\sigma_{\rm TD}^2,\\
\tau&=1,\\
B&=V_{\max}.
\end{align*}
Substituting these parameters into
\eqref{eq:meta-bound} gives
\begin{align*}
&\sum_{t=1}^T
\E\norm{\bar v_t-v_t^*}_\mu^2\\
&\quad\leq
4\inf_{\eta\in[1/T,1]}
\left\{
\frac{\norm{v_1-v_1^*}_\mu^2}
{\eta(1-\gamma)}
+\frac{\eta\sigma_{\rm TD}^2T}
{(1-\gamma)(2-\eta(1-\gamma))}
+\frac{W_T^V}{\eta^2(1-\gamma)^2}
\right\}\\
&\qquad+
8V_{\max}^2
\log(1+\lceil\log_2T\rceil),
\end{align*}
which proves the tracking guarantee.

Finally, we compute the total number
of generative-model calls. Since all
experts share the same sampled Bellman
operator, each TD update requires only
one call per state. There are $T-1$
update rounds, giving
\begin{align*}
(T-1)|\cS|
\end{align*}
calls in total.

By Proposition~\ref{prop:validation-td},
each validation vector requires an
expected $|\cS|/(1-\gamma)$ calls.
Validation is performed at all $T$
prediction rounds, so its expected
total sampling cost is
\begin{align*}
\frac{T|\cS|}{1-\gamma}.
\end{align*}
This establishes the sampling costs
stated in the corollary.
\end{proof}

\section{Linear recursions with two stepsizes}
\label{app:ttsa}

We extend the tracking analysis to linear stochastic
approximation with two coupled iterates and different
stepsizes. Finite-time analyses of linear two-timescale
stochastic approximation include
\citet{kaledin2020finite,haque2024tight,kwon2025constant}.
More recent work considers nonlinear recursions,
Markovian noise, nonexpansive maps, and operator-based
timescale separation
\citep{mahadevan2026convergence,sarkar2026nonlinear,
sarkar2026nonexpansive,carnevale2026timescale}.

Here, we focus on tracking when the equilibrium of
the coupled system changes over time. We consider
two settings. In the first, the full system is
contractive in a common quadratic norm. In the
second, the slow recursion evolves independently,
while the fast recursion depends on the slow iterate.
For this nested system, we derive separate tracking
bounds and relate the movement of the fast target
to changes in the system coefficients and the
slow recursion.

We use fixed inner-product norms on $\R^{d_x}$ and
$\R^{d_y}$, both denoted by $\norm{\cdot}$.
Operator norms are induced by the corresponding
domain and codomain norms. In particular,
$\norm{H_t}_{\op}$ is the induced norm of
$H_t:\R^{d_y}\to\R^{d_x}$.

\subsection{A stacked contraction bound}

We first consider two coupled linear recursions
with constant stepsizes $0<\beta\leq\alpha$:
\begin{align*}
x_{t+1}
&=x_t+\alpha\left(
A_{11,t}x_t+A_{12,t}y_t+c_{1,t}
+\xi_{t+1}\right),\\
y_{t+1}
&=y_t+\beta\left(
A_{21,t}x_t+A_{22,t}y_t+c_{2,t}
+\zeta_{t+1}\right),
\end{align*}
for $t=1,\ldots,T-1$, where
$x_t\in\R^{d_x}$ and $y_t\in\R^{d_y}$.
We assume that the coefficient sequences
and the initial iterates are deterministic.

To analyze the two recursions together,
define
\begin{align*}
z_t&=
\begin{bmatrix}
x_t\\y_t
\end{bmatrix},
\qquad
A_t=
\begin{bmatrix}
A_{11,t}&A_{12,t}\\
A_{21,t}&A_{22,t}
\end{bmatrix},\\
c_t&=
\begin{bmatrix}
c_{1,t}\\c_{2,t}
\end{bmatrix},
\qquad
D=
\begin{bmatrix}
\alpha I_{d_x}&0\\
0&\beta I_{d_y}
\end{bmatrix}.
\end{align*}
Writing
$\omega_{t+1}=(\xi_{t+1},\zeta_{t+1})$,
the coupled recursion becomes
\begin{align*}
z_{t+1}
=z_t+D(A_tz_t+c_t+\omega_{t+1}).
\end{align*}

We assume that there are a positive definite
matrix $Q=Q^\top\succ0$ and a constant
$\delta\in(0,1]$ such that
\begin{equation}
K_t^\top QK_t
\preceq(1-\delta)^2Q,
\qquad
K_t=I+DA_t,
\label{eq:common-lyap}
\end{equation}
for every $t$.

Let $\norm{z}_Q^2=z^\top Qz$.
Condition~\eqref{eq:common-lyap} implies
\begin{align*}
\norm{K_tz}_Q
\leq(1-\delta)\norm{z}_Q.
\end{align*}
Thus, every update contracts distances
in the same quadratic norm, even though
the coefficient matrices may change over
time. The condition also guarantees that
$A_t$ is invertible, as shown below.
We can therefore define the instantaneous
equilibrium of the coupled system as
\begin{align*}
z_t^*=-A_t^{-1}c_t.
\end{align*}

The following theorem bounds the cumulative
tracking error in terms of initialization,
noise, and movement of these equilibria.

\begin{theorem}[Tracking for a stacked contraction]
Suppose $\omega_{t+1}=(\xi_{t+1},\zeta_{t+1})$
is a martingale difference relative to the
joint filtration and
\begin{align*}
\E[\norm{D\omega_{t+1}}_Q^2\mid\cF_t]
\leq\bar\sigma^2.
\end{align*}
Under~\eqref{eq:common-lyap}, define
\begin{align*}
W_T^z
=\sum_{t=1}^{T-1}
\norm{z_{t+1}^*-z_t^*}_Q^2.
\end{align*}
Then
\begin{equation}
\sum_{t=1}^T\E\norm{z_t-z_t^*}_Q^2
\leq
\frac{\norm{z_1-z_1^*}_Q^2}{\delta}
+\frac{\bar\sigma^2T}{\delta(2-\delta)}
+\frac{W_T^z}{\delta^2}.
\label{eq:ttsa-bound}
\end{equation}
If
\begin{align*}
\E[\norm{\xi_{t+1}}_2^2\mid\cF_t]
&\leq\sigma_x^2,\\
\E[\norm{\zeta_{t+1}}_2^2\mid\cF_t]
&\leq\sigma_y^2,
\end{align*}
then one may take
\begin{align*}
\bar\sigma^2
=\lambda_{\max}(Q)
\left(\alpha^2\sigma_x^2
+\beta^2\sigma_y^2\right).
\end{align*}
\end{theorem}

\begin{proof}
We first verify that the instantaneous
equilibrium $z_t^*$ is well defined.
Suppose $A_tu=0$ for some vector $u$.
Since $K_t=I+DA_t$, we have $K_tu=u$.
On the other hand,
condition~\eqref{eq:common-lyap} gives
\begin{align*}
\norm{u}_Q
=\norm{K_tu}_Q
\leq(1-\delta)\norm{u}_Q.
\end{align*}
Because $\delta>0$, this is possible
only if $u=0$. Thus, $A_t$ is invertible,
and $z_t^*=-A_t^{-1}c_t$ is the unique
equilibrium satisfying
$A_tz_t^*+c_t=0$.

To analyze the tracking error, define
\begin{align*}
e_t&=z_t-z_t^*,\\
\Delta_t^z&=z_{t+1}^*-z_t^*.
\end{align*}
Using $A_tz_t^*+c_t=0$ in the coupled
recursion, we obtain
\begin{align*}
e_{t+1}
&=z_t+D(A_tz_t+c_t+\omega_{t+1})
-z_{t+1}^*\\
&=(I+DA_t)(z_t-z_t^*)
+D\omega_{t+1}-\Delta_t^z\\
&=K_te_t+D\omega_{t+1}-\Delta_t^z.
\end{align*}

We separate the error into its deterministic
and stochastic components by writing
$e_t=d_t+n_t$, where
\begin{align*}
d_1&=e_1,\qquad n_1=0,\\
d_{t+1}&=K_td_t-\Delta_t^z,\\
n_{t+1}&=K_tn_t+D\omega_{t+1}.
\end{align*}
Since the coefficients, equilibria, and
initial iterates are deterministic,
$d_t$ is deterministic. Moreover,
the martingale-difference assumption
implies $\E[n_t]=0$. Consequently,
\begin{align*}
\E\norm{e_t}_Q^2
=\norm{d_t}_Q^2+\E\norm{n_t}_Q^2.
\end{align*}

Let $\rho=1-\delta$. The common
contraction condition gives
$\norm{K_tz}_Q\leq\rho\norm{z}_Q$.
Applying Lemma~\ref{lem:deterministic-tracking}
in the $Q$ norm to the recursion for $d_t$,
we obtain
\begin{align*}
\sum_{t=1}^T\norm{d_t}_Q^2
&\leq
\frac{\norm{d_1}_Q^2}{1-\rho}
+\frac{1}{(1-\rho)^2}
\sum_{t=1}^{T-1}\norm{\Delta_t^z}_Q^2\\
&=
\frac{\norm{e_1}_Q^2}{\delta}
+\frac{W_T^z}{\delta^2}.
\end{align*}

We next bound the stochastic component.
For $t>s$, let
$\Phi(t,s)=K_{t-1}\cdots K_s$,
with $\Phi(s,s)=I$.
Unrolling the recursion for $n_t$ gives
\begin{align*}
n_t
=\sum_{s=1}^{t-1}
\Phi(t,s+1)D\omega_{s+1}.
\end{align*}
The matrices $\Phi(t,s+1)$ are
deterministic. Since the noise terms
form a martingale-difference sequence,
their cross terms vanish after taking
expectations. Hence,
\begin{align*}
\E\norm{n_t}_Q^2
&=\sum_{s=1}^{t-1}
\E\norm{\Phi(t,s+1)D\omega_{s+1}}_Q^2\\
&\leq
\bar\sigma^2
\sum_{s=1}^{t-1}\rho^{2(t-1-s)}\\
&\leq
\frac{\bar\sigma^2}{1-\rho^2},
\end{align*}
where we used the product contraction
bound from Lemma~\ref{lem:stable-products}.
Summing over $t$ gives
\begin{align*}
\sum_{t=1}^T\E\norm{n_t}_Q^2
&\leq
\frac{\bar\sigma^2(T-1)}{1-\rho^2}\\
&\leq
\frac{\bar\sigma^2T}{\delta(2-\delta)},
\end{align*}
where
$1-\rho^2=\delta(2-\delta)$.

Combining the deterministic and stochastic
bounds proves~\eqref{eq:ttsa-bound}.

Finally, we verify the stated choice of
$\bar\sigma^2$. Since
$\norm{w}_Q^2\leq
\lambda_{\max}(Q)\norm{w}_2^2$,
we have
\begin{align*}
&\E[\norm{D\omega_{t+1}}_Q^2\mid\cF_t]\\
&\quad\leq
\lambda_{\max}(Q)
\E[\norm{D\omega_{t+1}}_2^2\mid\cF_t]\\
&\quad=
\lambda_{\max}(Q)
\left(
\alpha^2\E[\norm{\xi_{t+1}}_2^2\mid\cF_t]
+\beta^2\E[\norm{\zeta_{t+1}}_2^2\mid\cF_t]
\right)\\
&\quad\leq
\lambda_{\max}(Q)
\left(\alpha^2\sigma_x^2
+\beta^2\sigma_y^2\right),
\end{align*}
which completes the proof.
\end{proof}

\subsection{Separate slow and fast bounds for a nested system}

We next consider a system in which the slow
recursion evolves independently, while the
fast recursion depends on the slow iterate.
Specifically, let
\begin{align*}
y_{t+1}
&=(1-\beta)y_t
+\beta(N_ty_t+g_t+\zeta_{t+1}),\\
x_{t+1}
&=(1-\alpha)x_t
+\alpha(M_tx_t+H_ty_t+f_t+\xi_{t+1}),
\end{align*}
where
\begin{align*}
\norm{N_t}_{\op}&\leq1-\kappa_y,\\
\norm{M_t}_{\op}&\leq1-\kappa_x,\\
0<\beta&\leq\alpha\leq1,
\end{align*}
for constants $\kappa_x,\kappa_y\in(0,1]$.
As before, all coefficient sequences and
initial iterates are deterministic.

The slow recursion has an instantaneous
fixed point
\begin{align*}
y_t^*=(I-N_t)^{-1}g_t.
\end{align*}
For the fast recursion, consider the
fixed point obtained by holding the
slow iterate at a given value $y$:
\begin{align*}
\Lambda_t(y)
=(I-M_t)^{-1}(H_ty+f_t).
\end{align*}
Thus, $y_t^*$ is the target of the slow
recursion, while $\Lambda_t(y_t)$ is the
target of the fast recursion at time $t$.

We measure the corresponding tracking
errors by
\begin{align*}
e_t^y&=y_t-y_t^*,\\
e_t^x&=x_t-\Lambda_t(y_t).
\end{align*}
Define their target variations as
\begin{align*}
W_T^y
&=\sum_{t=1}^{T-1}
\norm{y_{t+1}^*-y_t^*}^2,\\
W_T^\Lambda
&=\sum_{t=1}^{T-1}
\E\norm{\Lambda_{t+1}(y_{t+1})
-\Lambda_t(y_t)}^2.
\end{align*}

There is an important difference between
these two quantities. Since the coefficient
sequences are deterministic, the slow target
$y_t^*$ and its variation $W_T^y$ are also
deterministic. In contrast, the fast target
$\Lambda_t(y_t)$ depends on the random slow
iterate. Its movement therefore reflects
both changes in the system coefficients and
the stochastic evolution of $y_t$.

The following theorem gives separate bounds
for the two tracking errors.

\begin{theorem}[Separate slow and fast tracking bounds]
\label{thm:nested-ttsa}
Suppose $\xi_{t+1}$ and $\zeta_{t+1}$ are
martingale differences relative to the
joint filtration, with
\begin{align*}
\E[\norm{\xi_{t+1}}^2\mid\cF_t]
&\leq\sigma_x^2,\\
\E[\norm{\zeta_{t+1}}^2\mid\cF_t]
&\leq\sigma_y^2.
\end{align*}
Then the slow tracking error satisfies
\begin{align*}
\sum_{t=1}^T\E\norm{e_t^y}^2
&\leq
\frac{\norm{e_1^y}^2}{\beta\kappa_y}
+\frac{\beta\sigma_y^2T}
{\kappa_y(2-\beta\kappa_y)}
+\frac{W_T^y}{\beta^2\kappa_y^2}.
\end{align*}
The fast tracking error satisfies
\begin{align*}
\sum_{t=1}^T\E\norm{e_t^x}^2
&\leq
\frac{3\norm{e_1^x}^2}{\alpha\kappa_x}
+\frac{3\alpha\sigma_x^2T}
{\kappa_x(2-\alpha\kappa_x)}
+\frac{3W_T^\Lambda}{\alpha^2\kappa_x^2}.
\end{align*}
\end{theorem}

\begin{proof}
We begin with the slow recursion.
Since $y_t^*$ satisfies
$y_t^*=N_ty_t^*+g_t$, subtracting
$y_{t+1}^*$ from the slow update gives
\begin{align*}
e_{t+1}^y
&=(1-\beta)y_t
+\beta(N_ty_t+g_t+\zeta_{t+1})
-y_{t+1}^*\\
&=((1-\beta)I+\beta N_t)e_t^y
+\beta\zeta_{t+1}
-(y_{t+1}^*-y_t^*).
\end{align*}
The linear coefficient satisfies
\begin{align*}
\norm{(1-\beta)I+\beta N_t}_{\op}
&\leq(1-\beta)+\beta(1-\kappa_y)\\
&=1-\beta\kappa_y.
\end{align*}
Thus, the slow recursion has the same
form as the affine tracking iteration
studied in Theorem~\ref{thm:upper},
with stepsize $\beta$, contraction gap
$\kappa_y$, and martingale-difference
noise $\zeta_{t+1}$. Applying that theorem
with $\tau=1$ gives
\begin{align*}
\sum_{t=1}^T\E\norm{e_t^y}^2
&\leq
\frac{\norm{e_1^y}^2}{\beta\kappa_y}
+\frac{\beta\sigma_y^2T}
{\kappa_y(2-\beta\kappa_y)}
+\frac{W_T^y}{\beta^2\kappa_y^2},
\end{align*}
which proves the first bound.

We now turn to the fast recursion.
By the definition of $\Lambda_t$,
\begin{align*}
\Lambda_t(y_t)
=M_t\Lambda_t(y_t)+H_ty_t+f_t.
\end{align*}
Subtracting $\Lambda_{t+1}(y_{t+1})$
from the fast update and using this
identity gives
\begin{align*}
e_{t+1}^x
&=((1-\alpha)I+\alpha M_t)e_t^x
+\alpha\xi_{t+1}\\
&\quad-
\left(
\Lambda_{t+1}(y_{t+1})-\Lambda_t(y_t)
\right).
\end{align*}

For convenience, define
\begin{align*}
K_t^x&=(1-\alpha)I+\alpha M_t,\\
\Delta_t^\Lambda
&=\Lambda_{t+1}(y_{t+1})-\Lambda_t(y_t).
\end{align*}
The error recursion becomes
\begin{align*}
e_{t+1}^x
=K_t^xe_t^x+\alpha\xi_{t+1}
-\Delta_t^\Lambda.
\end{align*}
Moreover,
\begin{align*}
\norm{K_t^x}_{\op}
&\leq1-\alpha\kappa_x
=:\rho_x<1.
\end{align*}

Unlike the slow recursion, the target
increment $\Delta_t^\Lambda$ is random.
It may also be correlated with
$\xi_{t+1}$ because both recursions
evolve using observations from the
same time period. We therefore cannot
directly use the deterministic--stochastic
decomposition from
Theorem~\ref{thm:upper}.
Instead, we bound the contributions of
initialization, observation noise, and
target movement separately.

For $t>s$, define
\begin{align*}
\Phi_x(t,s)
=K_{t-1}^x\cdots K_s^x,
\qquad
\Phi_x(s,s)=I.
\end{align*}
Unrolling the fast error recursion gives
\begin{align*}
e_t^x
&=\Phi_x(t,1)e_1^x\\
&\quad+
\alpha\sum_{s=1}^{t-1}
\Phi_x(t,s+1)\xi_{s+1}\\
&\quad-
\sum_{s=1}^{t-1}
\Phi_x(t,s+1)\Delta_s^\Lambda.
\end{align*}

Using
$\norm{a+b+c}^2
\leq3(\norm{a}^2+\norm{b}^2+\norm{c}^2)$,
we can bound the expected squared error
by three corresponding terms.

For the initial error, the stable-product
bound in Lemma~\ref{lem:stable-products}
gives
\begin{align*}
\sum_{t=1}^T
\norm{\Phi_x(t,1)e_1^x}^2
&\leq
\frac{\norm{e_1^x}^2}{1-\rho_x^2}\\
&\leq
\frac{\norm{e_1^x}^2}{1-\rho_x}\\
&=
\frac{\norm{e_1^x}^2}{\alpha\kappa_x}.
\end{align*}

For the observation noise, the
martingale-difference assumption and
Lemma~\ref{lem:noise-accumulation},
applied with $\tau=1$, give
\begin{align*}
&\alpha^2\sum_{t=2}^T
\E\norm{
\sum_{s=1}^{t-1}
\Phi_x(t,s+1)\xi_{s+1}
}^2\\
&\quad\leq
\frac{\alpha^2\sigma_x^2(T-1)}
{1-\rho_x^2}\\
&\quad\leq
\frac{\alpha\sigma_x^2T}
{\kappa_x(2-\alpha\kappa_x)}.
\end{align*}

Finally, consider the contribution of
the moving fast target.
Lemma~\ref{lem:stable-convolution} applies
to the sequence
$\Delta_1^\Lambda,\ldots,
\Delta_{T-1}^\Lambda$ for every realization
of the slow recursion. Taking expectations
in the resulting bound gives
\begin{align*}
&\sum_{t=2}^T
\E\norm{
\sum_{s=1}^{t-1}
\Phi_x(t,s+1)\Delta_s^\Lambda
}^2\\
&\quad\leq
\frac{1}{(1-\rho_x)^2}
\sum_{s=1}^{T-1}
\E\norm{\Delta_s^\Lambda}^2\\
&\quad=
\frac{W_T^\Lambda}{\alpha^2\kappa_x^2}.
\end{align*}

Combining these three bounds yields
\begin{align*}
\sum_{t=1}^T\E\norm{e_t^x}^2
&\leq
\frac{3\norm{e_1^x}^2}{\alpha\kappa_x}
+\frac{3\alpha\sigma_x^2T}
{\kappa_x(2-\alpha\kappa_x)}\\
&\quad+
\frac{3W_T^\Lambda}{\alpha^2\kappa_x^2},
\end{align*}
as claimed.

The argument does not require independence
between the two recursions. In particular,
the possible dependence between the
fast observation noise and the movement
of its target is handled by bounding
their contributions separately.
\end{proof}

The fast tracking bound depends on
$W_T^\Lambda$, which includes the
random movement induced by the slow iterate.
We next relate this quantity to the
slow tracking error and changes in the
fast fixed-point map.

For the following results, we assume
that the unprojected slow iterates
$y_1,\ldots,y_T$ remain in a set
$\cY$ almost surely. This condition
may follow from a forward-invariant set
or a separate boundedness argument.
It does not follow from a second-moment
bound on the noise alone.

\begin{proposition}[Bounding movement of the fast fixed-point map]
\label{prop:wlambda}
Under the assumptions of
Theorem~\ref{thm:nested-ttsa},
suppose that each map $\Lambda_t$
is $L_\Lambda$-Lipschitz on a set
$\cY$ containing $y_1,\ldots,y_T$
almost surely.

Define
\begin{align*}
d_t^\Lambda
&=\sup_{y\in\cY}
\norm{\Lambda_{t+1}(y)-\Lambda_t(y)},\\
\Omega_T^\Lambda
&=\sum_{t=1}^{T-1}(d_t^\Lambda)^2,\\
\rho_y
&=\sup_t\norm{N_t}_{\op}
\leq1-\kappa_y.
\end{align*}
Then
\begin{align*}
W_T^\Lambda
&\leq
2\Omega_T^\Lambda\\
&\quad+
2L_\Lambda^2\beta^2(1+\rho_y)^2
\sum_{t=1}^{T-1}\E\norm{e_t^y}^2\\
&\quad+
2L_\Lambda^2\beta^2\sigma_y^2T.
\end{align*}
\end{proposition}

\begin{proof}
We begin by separating the movement of
the fast target into two parts. Adding
and subtracting $\Lambda_{t+1}(y_t)$
gives
\begin{align*}
&\Lambda_{t+1}(y_{t+1})-\Lambda_t(y_t)\\
&\quad=
\left(
\Lambda_{t+1}(y_{t+1})
-\Lambda_{t+1}(y_t)
\right)\\
&\qquad+
\left(
\Lambda_{t+1}(y_t)-\Lambda_t(y_t)
\right).
\end{align*}
The first term measures the change in
the fast target caused by the movement
of the slow iterate. The second measures
the change in the fast fixed-point map
when the slow iterate is held fixed.

By the Lipschitz assumption and the
definition of $d_t^\Lambda$,
\begin{align*}
&\norm{\Lambda_{t+1}(y_{t+1})
-\Lambda_t(y_t)}\\
&\quad\leq
L_\Lambda\norm{y_{t+1}-y_t}
+d_t^\Lambda.
\end{align*}
Using $(a+b)^2\leq2a^2+2b^2$
and taking expectations gives
\begin{align*}
&\E\norm{\Lambda_{t+1}(y_{t+1})
-\Lambda_t(y_t)}^2\\
&\quad\leq
2L_\Lambda^2
\E\norm{y_{t+1}-y_t}^2
+2(d_t^\Lambda)^2.
\end{align*}

It remains to bound the movement of
the slow iterate. Recall that
$y_t^*=N_ty_t^*+g_t$.
Subtracting $y_t$ from the slow
update gives
\begin{align*}
y_{t+1}-y_t
&=\beta\left(
(N_t-I)y_t+g_t+\zeta_{t+1}
\right)\\
&=\beta\left(
(N_t-I)e_t^y+\zeta_{t+1}
\right).
\end{align*}

Since $e_t^y$ is measurable with
respect to $\cF_t$ and
$\E[\zeta_{t+1}\mid\cF_t]=0$,
the cross term vanishes after
taking expectations. Hence,
\begin{align*}
\E\norm{y_{t+1}-y_t}^2
&=\beta^2\E\norm{(N_t-I)e_t^y}^2\\
&\quad+
\beta^2\E\norm{\zeta_{t+1}}^2.
\end{align*}
Using
$\norm{N_t-I}_{\op}
\leq1+\norm{N_t}_{\op}
\leq1+\rho_y$,
we obtain
\begin{align*}
\E\norm{y_{t+1}-y_t}^2
&\leq
\beta^2(1+\rho_y)^2
\E\norm{e_t^y}^2
+\beta^2\sigma_y^2.
\end{align*}

Substituting this bound into the
earlier inequality and summing over
$t=1,\ldots,T-1$ yields
\begin{align*}
W_T^\Lambda
&\leq
2\Omega_T^\Lambda\\
&\quad+
2L_\Lambda^2\beta^2(1+\rho_y)^2
\sum_{t=1}^{T-1}\E\norm{e_t^y}^2\\
&\quad+
2L_\Lambda^2\beta^2\sigma_y^2(T-1)\\
&\leq
2\Omega_T^\Lambda\\
&\quad+
2L_\Lambda^2\beta^2(1+\rho_y)^2
\sum_{t=1}^{T-1}\E\norm{e_t^y}^2\\
&\quad+
2L_\Lambda^2\beta^2\sigma_y^2T,
\end{align*}
which proves the claim.
\end{proof}

The preceding proposition separates the
movement of the fast target into changes
in the fixed-point map, the slow tracking
error, and the slow observation noise.
The remaining quantity
$\Omega_T^\Lambda$ measures how much
the fast fixed-point map changes over
time when its argument is held fixed.

We next bound this quantity directly
using changes in the coefficients
$M_t$, $H_t$, and $f_t$.

\begin{proposition}[Coefficient-based bound]
Assume
\begin{align*}
\sup_{y\in\cY}\norm{y}&\leq R_y,\\
\bar H:=\sup_t\norm{H_t}_{\op}&<\infty,\\
\bar f:=\sup_t\norm{f_t}&<\infty.
\end{align*}
Define
\begin{align*}
C_\Lambda
&=\frac{\bar H R_y+\bar f}{\kappa_x},\\
W_T^M
&=\sum_{t=1}^{T-1}
\norm{M_{t+1}-M_t}_{\op}^2,\\
W_T^H
&=\sum_{t=1}^{T-1}
\norm{H_{t+1}-H_t}_{\op}^2,\\
W_T^f
&=\sum_{t=1}^{T-1}
\norm{f_{t+1}-f_t}^2.
\end{align*}
Then each $\Lambda_t$ is Lipschitz
on $\cY$ with common constant
\begin{align*}
L_\Lambda=\frac{\bar H}{\kappa_x},
\end{align*}
and
\begin{align*}
\Omega_T^\Lambda
\leq
\frac{3}{\kappa_x^2}
\left(
R_y^2W_T^H
+W_T^f
+C_\Lambda^2W_T^M
\right).
\end{align*}
Together with
Proposition~\ref{prop:wlambda}
and Theorem~\ref{thm:nested-ttsa},
this gives a fast tracking bound
in terms of the coefficient variations
and the slow tracking error.
\end{proposition}

\begin{proof}
Recall that
\begin{align*}
\Lambda_t(y)
=(I-M_t)^{-1}(H_ty+f_t).
\end{align*}
Since
$\norm{M_t}_{\op}\leq1-\kappa_x<1$,
the Neumann series gives
\begin{align*}
\norm{(I-M_t)^{-1}}_{\op}
&\leq
\sum_{k=0}^\infty\norm{M_t}_{\op}^k\\
&\leq\frac1{\kappa_x}.
\end{align*}

We first verify the Lipschitz condition.
For any $y,y'\in\cY$,
\begin{align*}
\norm{\Lambda_t(y)-\Lambda_t(y')}
&=\norm{(I-M_t)^{-1}H_t(y-y')}\\
&\leq
\frac{\norm{H_t}_{\op}}{\kappa_x}
\norm{y-y'}\\
&\leq
\frac{\bar H}{\kappa_x}\norm{y-y'}.
\end{align*}
Thus, we may take
$L_\Lambda=\bar H/\kappa_x$.

The same resolvent bound gives,
for every $y\in\cY$,
\begin{align*}
\norm{\Lambda_t(y)}
&\leq
\frac1{\kappa_x}
\left(
\norm{H_t}_{\op}\norm{y}
+\norm{f_t}
\right)\\
&\leq
\frac{\bar H R_y+\bar f}{\kappa_x}\\
&=C_\Lambda.
\end{align*}

We next bound the change in
$\Lambda_t(y)$ between consecutive
rounds. From the fixed-point identities
\begin{align*}
(I-M_t)\Lambda_t(y)&=H_ty+f_t,\\
(I-M_{t+1})\Lambda_{t+1}(y)
&=H_{t+1}y+f_{t+1},
\end{align*}
we obtain, after subtraction,
\begin{align*}
&(I-M_{t+1})
\left(\Lambda_{t+1}(y)-\Lambda_t(y)\right)\\
&\quad=
(H_{t+1}-H_t)y+(f_{t+1}-f_t)\\
&\qquad+
(M_{t+1}-M_t)\Lambda_t(y).
\end{align*}

Multiplying by $(I-M_{t+1})^{-1}$
and taking norms, we have
\begin{align*}
&\norm{\Lambda_{t+1}(y)-\Lambda_t(y)}\\
&\quad\leq
\frac1{\kappa_x}
\left(
\norm{H_{t+1}-H_t}_{\op}\norm{y}
+\norm{f_{t+1}-f_t}\right.\\
&\hspace{3.2cm}\left.
+\norm{M_{t+1}-M_t}_{\op}
\norm{\Lambda_t(y)}
\right).
\end{align*}
Using $\norm{y}\leq R_y$ and
$\norm{\Lambda_t(y)}\leq C_\Lambda$,
then taking the supremum over
$y\in\cY$, gives
\begin{align*}
d_t^\Lambda
&\leq
\frac1{\kappa_x}
\left(
R_y\norm{H_{t+1}-H_t}_{\op}
+\norm{f_{t+1}-f_t}\right.\\
&\hspace{2.9cm}\left.
+C_\Lambda\norm{M_{t+1}-M_t}_{\op}
\right).
\end{align*}

Finally, applying
$(a+b+c)^2\leq3(a^2+b^2+c^2)$,
we obtain
\begin{align*}
(d_t^\Lambda)^2
&\leq
\frac3{\kappa_x^2}
\left(
R_y^2\norm{H_{t+1}-H_t}_{\op}^2
+\norm{f_{t+1}-f_t}^2\right.\\
&\hspace{3cm}\left.
+C_\Lambda^2
\norm{M_{t+1}-M_t}_{\op}^2
\right).
\end{align*}
Summing over $t=1,\ldots,T-1$
and using the definitions of
$W_T^M$, $W_T^H$, and $W_T^f$
proves
\begin{align*}
\Omega_T^\Lambda
&=\sum_{t=1}^{T-1}(d_t^\Lambda)^2\\
&\leq
\frac3{\kappa_x^2}
\left(
R_y^2W_T^H
+W_T^f
+C_\Lambda^2W_T^M
\right),
\end{align*}
as claimed.
\end{proof}

\section{Comparison with prior work}
\label{app:comparison}

Table~\ref{tab:comparison} compares our results with
selected work on tracking time-varying targets.
These results differ in their assumptions, observation
models, and measures of target variation, so their
bounds are not always directly comparable. We
summarize these differences to clarify how our
results relate to the existing literature.

\begin{table}[H]
\centering
\caption{Comparison of selected tracking and estimation
guarantees. Numerical constants are suppressed, and
$\widetilde O$ hides logarithmic factors.}
\label{tab:comparison}
\footnotesize
\setlength{\tabcolsep}{3pt}
\renewcommand{\arraystretch}{1.15}
\begin{tabularx}{\textwidth}{
@{}L{1.9cm}L{3.35cm}L{2.35cm}
>{\raggedright\arraybackslash}X@{}}
\toprule
Work & Model and observations & Target variation
& Selected guarantee \\
\midrule
\citet{hadiji2026tracking}, Theorem 2.1
& Contractive updates with exact operator evaluations
& Sum of squared target increments
& Cumulative squared tracking error with
initialization and target-movement terms \\

\citet{bastianello2024stochastic}, Propositions 1 and 3
& General time-varying contractive operators
with possibly biased observation errors
& Bound on target movement per round
& Expected-distance tracking under
second-moment-bounded additive errors;
high-probability guarantees under
sub-Weibull error assumptions and
additional conditions on the random
update operators \\

\citet{cutler2023distributional}, Theorem 3
& Strongly convex optimization with
unbiased stochastic gradients
& Uniform bound on expected squared
minimizer increments
& Pointwise expected squared tracking error
for $\eta\leq1/(2L)$, with initialization,
noise, and movement terms; the latter two
scale as
$\eta\sigma^2/m+\Delta^2/(m^2\eta^2)$ \\

\citet{baby2019online}, Section~3.5
& Scalar target observations with independent
sub-Gaussian noise; known horizon and noise parameter
& $\norm{D\theta}_2^2\leq W$
& Adaptive high-probability squared-error bound
$\widetilde O(T^{2/3}W^{1/3}\nu^{4/3}
+U^2+W+\nu^2)$ \\

This paper,
Theorems~\ref{thm:upper} and~\ref{thm:meta}
& Deterministic affine contractions with centered
noise satisfying~\eqref{eq:integrated-covariance}
& Sum of squared fixed-point increments, $W_T$
& Cumulative mean-square tracking bound;
stepsize adaptation using additional bounded,
unbiased target observations \\
\bottomrule
\end{tabularx}

\vspace{3pt}
\parbox{\textwidth}{\footnotesize
In the optimization row, $m$ and $L$ denote
the strong-convexity and smoothness parameters,
respectively, and $\Delta^2$ bounds the expected
squared minimizer increment at each round. In the forecasting
row, $\nu^2$ denotes the noise proxy and
$\abs{\theta_1}\leq U$.
}
\end{table}

Our tracking bound builds on the deterministic
contraction result of \citet{hadiji2026tracking}
by accounting for observation noise and temporal
correlation. Compared with stochastic tracking
results for more general operators, our analysis
uses the affine structure to obtain a cumulative
squared-error bound in terms of the quadratic
variation $W_T$. For stepsize adaptation, our
approach requires additional unbiased observations
of the target, whereas the scalar forecasting
method of \citet{baby2019online} adapts without
a separate validation stream. These differences
reflect both the scope of our guarantees and
the assumptions needed to obtain them.

\end{document}

%% file: math_commands.tex
\usepackage{amsmath,amsfonts,bm}

\def\eqref#1{equation~\ref{#1}}

\def\1{\bm{1}}

\DeclareMathAlphabet{\mathsfit}{\encodingdefault}{\sfdefault}{m}{sl}
\SetMathAlphabet{\mathsfit}{bold}{\encodingdefault}{\sfdefault}{bx}{n}

\newcommand{\E}{\mathbb{E}}

\newcommand{\R}{\mathbb{R}}

